\PassOptionsToPackage{table,dvipsnames,svgnames}{xcolor}
\documentclass{article} 
\usepackage{iclr2027_conference,times}

\usepackage{amsmath,amsfonts,bm}

\def\eqref#1{equation~\ref{#1}}

\def\1{\bm{1}}

\DeclareMathAlphabet{\mathsfit}{\encodingdefault}{\sfdefault}{m}{sl}
\SetMathAlphabet{\mathsfit}{bold}{\encodingdefault}{\sfdefault}{bx}{n}

\DeclareMathOperator*{\argmax}{arg\,max}

\usepackage[pagebackref,breaklinks,colorlinks,citecolor=MidnightBlue,linkcolor=BrickRed]{hyperref}
\usepackage{url}
\usepackage{graphicx}
\usepackage{multirow}
\usepackage{multicol}
\usepackage[table]{xcolor}
\usepackage{booktabs}

\usepackage[table,dvipsnames,svgnames]{xcolor}
\usepackage{algpseudocode}
\usepackage{graphicx}
\usepackage{booktabs}
\usepackage{array}

\usepackage{wrapfig}
\usepackage{colortbl}
\usepackage{makecell}
\usepackage{pifont}

\usepackage{multirow}

\usepackage{algorithm,algcompatible,amsmath,}
\usepackage{makecell}

\usepackage{listings}

\usepackage[utf8]{inputenc} 
\usepackage[T1]{fontenc}    
\usepackage{hyperref}       
\usepackage{url}            
\usepackage{amsfonts}       
\usepackage{nicefrac}       
\usepackage{microtype}      
\usepackage{enumitem}

\algdef{SE}[SUBALG]{Indent}{EndIndent}{}{\algorithmicend\ }%
\algnewcommand\INPUT{\item[\textbf{Input:}]}%
\algnewcommand\OUTPUT{\item[\textbf{Output:}]}%
\algnewcommand\PARAM{\item[\textbf{Optimizable Parameter:}]}%
\algtext*{Indent}
\algtext*{EndIndent}

\def\methodname{ActiveLang}

\definecolor{tealblue}{rgb}{0.212,0.459, 0.533}

\definecolor{modeoffline}{HTML}{EBEBEB}
\definecolor{modeonline}{HTML}{DCE9FA}
\definecolor{modeactive}{HTML}{E6DCF5}
\definecolor{vocnone}{HTML}{F4F4F4}
\definecolor{vocobject}{HTML}{FFF2CC}
\definecolor{vocclosed}{HTML}{FCE0D0}
\definecolor{vocopen}{HTML}{D8F0DC}
\newcommand{\mOff}{\cellcolor{modeoffline}Offline}
\newcommand{\mOn}{\cellcolor{modeonline}Online}
\newcommand{\mAct}{\cellcolor{modeactive}Active}
\newcommand{\vNone}{\cellcolor{vocnone}--}
\newcommand{\vObj}{\cellcolor{vocobject}Object}
\newcommand{\vClosed}{\cellcolor{vocclosed}Closed}
\newcommand{\vOpen}{\cellcolor{vocopen}Open}
\newcommand{\vObjOpen}{\cellcolor{vocopen}Object, Open}

\title{ActiveLang: Active Open-Vocabulary 3D Mapping with Semantic-Uncertainty-Guided Exploration}

\author{
Liyan Chen 
\thanks{Corresponding author (lchen39@stevens.edu)}
\textsuperscript{1},
Hairong Yin 
\textsuperscript{2},
Huangying Zhan 
\textsuperscript{3},
Yi Xu 
\textsuperscript{3},
Raymond A. Yeh
\textsuperscript{2},
Philippos Mordohai 
\textsuperscript{1} \\
\normalsize{
\textsuperscript{1}
    Stevens Institute of Technology
\quad
\textsuperscript{2}
    Purdue University
\quad
\textsuperscript{3}
   Goertek Alpha Labs}
   }

\iclrfinalcopy 
\begin{document}

\maketitle
\vspace{-0.35cm}
\begin{figure}[H]
    \centering
    \vspace{-20pt}
    \includegraphics[width=0.8\textwidth]{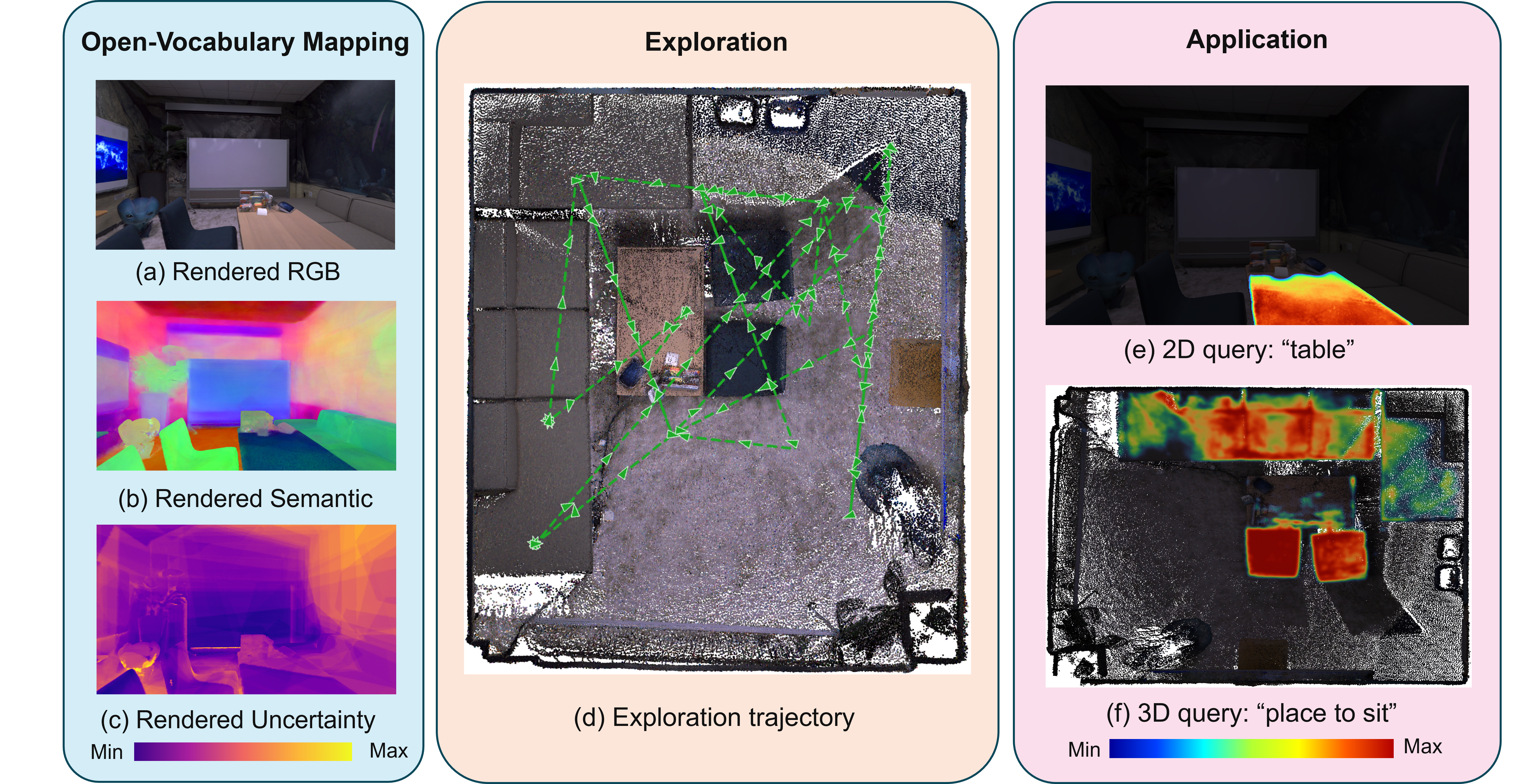}
    \vspace{-0.25cm}
    \caption{\textbf{ActiveLang integrates open-vocabulary 3D mapping with autonomous exploration.}
\textbf{Left:} ActiveLang reconstructs semantic RGB-D  Gaussian maps. The rendered semantic maps show coherent regions that largely follow true semantic boundaries, while learned uncertainty guides exploration.
\textbf{Middle:} The autonomous trajectory overlaid on the reconstructed 3D map illustrates that exploration achieves scene coverage.
\textbf{Right:} The semantic Gaussians support open-vocabulary queries in individual novel views (``table'') and directly on the 3D point cloud (``place to sit'').
}
    \label{fig:teaser}
    \vspace{-16pt}
\end{figure}

\begin{abstract}
As robots increasingly assist humans with diverse tasks, they need both geometric and semantic understanding of their surroundings. Moreover, robots often operate in unfamiliar environments and take on new tasks without knowing the relevant concepts ahead of time. This motivates language-annotated 3D maps that support open-vocabulary scene understanding and human–robot interaction. We introduce ActiveLang, an autonomous system for active open-vocabulary 3D mapping with semantic-uncertainty-guided exploration. ActiveLang performs online language-feature adaptation on a compact dual-Gaussian representation to jointly reconstruct scene geometry, appearance, and open-vocabulary semantics with modest memory overhead. Its planner efficiently selects informative viewpoints, enabling effective mapping with fewer observations and lower computational cost. Experiments on Replica and ScanNet++ demonstrate substantial improvements in 2D and 3D open-vocabulary segmentation over both online and offline baselines, highlighting that actively exploring scenes builds language-annotated 3D maps more efficiently. 
\end{abstract}

\section{Introduction}\label{sec:intro}

As embodied AI advances, robots are increasingly deployed in diverse environments and tasked with a growing range of activities. Operating effectively in these settings requires an accurate understanding of both the geometry and semantics of their surroundings. However, conventional closed-vocabulary semantic mapping methods \citep{li2024sgs}, rely on {\it predefined sets of labels}, which may not cover the concepts needed for unfamiliar environments or new tasks, or align with the terms used by humans. 
To facilitate natural human–robot interaction and support 
robots operating in unfamiliar environments, semantic maps that are directly queryable through natural language are needed. Language-annotated 3D maps meet this requirement by associating spatial scene representations with language features
\citep{qin2024langsplat}.

Creating such maps requires multiple observations from the robot's sensors. Reconstruction can be performed offline in batch mode \citep{qin2024langsplat} or online on the data stream \citep{katragadda2025_onlinelang}. We refer to mapping methods as \textit{passive} when the trajectory of the sensor is imposed externally, e.g. when the robot is driven by a human operator, and as \textit{active} when the robot determines the trajectory \textit{autonomously} \citep{ahmed2023active}. Active mapping, therefore, must operate on streams of data that are dynamically determined as information about the environments is revealed \citep{lluvia2021active,placed2023survey}. Determining the trajectory autonomously relieves human operators from the time and effort needed to drive the robot, and possibly also from the effort to familiarize themselves with characteristics of the environment and the robot. If the exploration strategy is effective, active mapping is capable of generating a map of the same quality while limiting acquisition effort compared to passive mapping. 

In this paper, we address \textbf{active open-vocabulary 3D mapping}  at the level of pixels and 3D primitives; see Fig.~\ref{fig:teaser}. This is a much more challenging than tasks addressed by prior works, none of which tackle active mapping of all objects and surfaces with an open vocabulary.
Our method aims for dense semantic representations that are not restricted to foreground objects, without knowing a priori the number or type of visual categories that are present in the environment. To accomplish this goal, we must overcome the following challenges.

First, open-vocabulary mapping suffers from embedding misalignment across modalities.
Unlike in closed-vocabulary settings, where the mapping between language and visual semantic embeddings is given, open-vocabulary semantics are characterized by ambiguity due to a plethora of causes, including ambiguous granularity, e.g. \textit{table} and \textit{desk} may be treated as the same or different categories, and multi-modality of the embeddings, e.g. the embeddings of legs, wheels and backs of different \textit{chairs} may be similar in language space and different in visual space. 
High-dimensional language features also impose substantial storage, optimization, and rendering costs, motivating feature compression \citep{qin2024langsplat,katragadda2025_onlinelang}. During active exploration, however, new concepts appear progressively and observations are unevenly distributed across the scene. Learning a compact representation from this evolving stream can lead to bias in favor of frequently observed content and may blur semantic distinctions, making efficient compression and semantic preservation difficult to achieve jointly

Second, selecting observations that jointly improve geometric and semantic map quality is challenging. A viewpoint that reveals previously unseen surfaces may provide weak or inconsistent language features, while revisiting a geometrically accurate region may help resolve semantic ambiguity without expanding coverage.  
Designing an exploration strategy that strikes a good balance between these two criteria is hard.
Several recent approaches for open-vocabulary exploration and
navigation rely on frontiers \citep{alama2025rayfronts,padilla2026openfrontier,schperberg2026roboatlas}, possibly to bypass the formulation of an uncertainty-driven exploration criterion.

To address these challenges, we introduce \methodname{}, a
closed-loop system for active open-vocabulary 3D mapping. 
Our main contributions are:
\vspace{-3pt}
\begin{itemize}[noitemsep, topsep=0pt, leftmargin=12pt]

\item \textbf{A closed-loop framework for dense language mapping
and exploration.}
\methodname{} incrementally reconstructs geometry, appearance, and language features while autonomously selecting its next observations. By coupling map updates with viewpoint selection, it builds language-queryable maps of the scene.

\item \textbf{Compact, semantic-preserving representation.}
We adapt a language-feature autoencoder as observations arrive
to reduce the cost of dense semantic representation.
We introduce a prototype-based adaptation objective that balances supervision across feature modes while preserving their separation during compression, reducing bias towards frequently observed content and semantic confusion.

\item \textbf{Exploration guided by geometric coverage and
semantic uncertainty.}
Our semantic loss jointly considers the semantics rendered by the Gaussians and their predicted uncertainty, and uses the latter as a proxy for the potential informativeness of additional observations.
Our next-best-view criterion combines this signal with geometric coverage and travel cost to balance exploration of unseen surfaces with refinement of uncertain semantics.

\end{itemize}

\section{Related Work}\label{sec:rel_work}
Most relevant to our work are methods that address active and semantic mapping, separately or jointly, using Neural Radiance Fields (NeRF) \citep{mildenhall2020nerf,yao2026neural_survey} and Gaussian Splatting (GS) \citep{kerbl20233dgs,chen2026survey} to represent the map.
Surveys on semantic mapping and on using NeRF for it were published by \citet{raychaudhuri2025semantic} and \citet{nguyen2024semantically}, respectively. We classify prior work along two axes: 
\textit{mapping mode}, which may be offline when all data are available simultaneously, online when data are processed sequentially, and active when the robot determines the next data to be acquired; and type of \textit{semantic vocabulary}, which may be foreground objects only, closed or open.

\begin{wraptable}{r}{0.45\textwidth}{
\centering
\vspace{-20pt}
\caption{Comparison with representative passive and active mapping methods. All methods address appearance and geometric mapping. Semantic vocabulary is none (--), object-level, closed-set, or open-set. }
\vspace{5pt}
\label{tab:method_comparison}
\resizebox{0.45\textwidth}{!}{
\begin{tabular}{lcc}
\toprule
Method & Mapping Mode & Semantic Vocabulary \\
\midrule
GSNeRF                    & \mOff & \vClosed  \\
LangSplat                 & \mOff & \vOpen    \\
OpenGaussian              & \mOff & \vOpen    \\
\addlinespace[2pt]
SplaTAM                   & \mOn  & \vNone    \\
SGS-SLAM                  & \mOn  & \vClosed  \\
OVI-MAP                   & \mOn  & \vObjOpen \\
Online Language Splatting & \mOn  & \vOpen    \\
\addlinespace[2pt]
ActiveGAMER               & \mAct & \vNone    \\
AutoNeRF                  & \mAct & \vObj     \\
ActiveSGM                 & \mAct & \vClosed  \\
\midrule
\textbf{ActiveLang (Ours)} & \mAct & \vOpen   \\
\bottomrule
\end{tabular}
}
}
\vspace{-15pt}
\end{wraptable}

Table~\ref{tab:method_comparison} shows representative methods from each category. For example, GSNeRF \citep{chou2024gsnerf} is an offline, closed-vocabulary approach.
Representative semantic SLAM approaches with GS mapping backbones include SGS-SLAM \citep{li2024sgs}, NIDS-SLAM \citep{haghighi2023nids}, SNI-SLAM \citep{zhu2024sni}, DNS-SLAM \citep{li2024dns}, SemGauss-SLAM \citep{zhu2024semgauss}  and Hier-SLAM \citep{li2025hier}. 

\textbf{Active Mapping.} 
\citet{lluvia2021active} and \citet{placed2023survey} surveyed the  literature.
Recently, NeRF-based representations have been used for path planning \citep{adamkiewicz2022vision} and next-best-view selection \citep{jin2023neu,lee2022uncertainty, pan2022activenerf,ran2023neurar,xue2024neural,yan2023activeimplicit}, while hybrid implicit-explicit representations have also been proposed \citep{zhan2022activermap,feng2024naruto,kuang2024active} to reduce computational cost.
GS-based uncertainty predictors are faster and more scalable. They are often combined with additional representations in approaches such as AG-SLAM \citep{jiang2024agslam}, RT-GuIDE \citep{tao2025rtguide}, ActiveSplat \citep{li2025activesplat}, ActiveGS \citep{jin2025activegs}, 
ActiveGAMER \citep{chen2025activegamer} 
and NextBestPath \citep{li2025nextbestpath}. AREA3D \citep{xu2026area3d} relies on semantic uncertainty for exploration, but it only maps RGB-D.

\textbf{Active Closed-vocabulary Mapping.} These methods  determine sensor trajectory autonomously, but do not face all challenges of Section~\ref{sec:intro} since labels do not need to be mapped across modalities (from language to images). \citet{li2026next} 
select among discrete views for semantic mapping, neglecting travel cost.
Other methods focus on objects, but not surfaces 
\citep{gervet2023navigating,georgakis2022learning,ramakrishnan2022poni}.
\citet{zhang2024active} guide exploration via semantic mutual information and properties of the SLAM pose graph, but their approach is limited to 8 classes. 
AutoNeRF\citep{marza2024autonerf} added semantics to the NeRF backbone, but is limited to 2D motion and 15 classes.
ActiveSGM \citep{chen2025understanding} labels both the foreground and background, handling over 100 labels while exploring the scene guided by geometric and semantic uncertainty.

\textbf{Open-vocabulary Mapping.} Early methods were offline, starting with LERF \cite{kerr2023lerf}, which  endowed the volumetric rendering of NeRF with semantic features, supervised by CLIP.
OpenNeRF  \citet{engelmann2024opennerf} uses pixel-wise VLM
features, instead of LERF's global CLIP features, resulting in a lighter architecture.
The most representative offline open-vocabulary semantic mapping method is LangSplat \citep{qin2024langsplat}. Similar to our work, LangSplat trains a scene-specific autoencoder to capture  distinctions between the categories that are present. LangSplat, however, trains the autoencoder offline on all frames, while we have to adapt it online. It also relies on SAM \cite{kirillov2023segment}, which we do not use.
OpenGaussian \citep{wu2024opengaussian} advocates for reasoning on the 3D splats rather than on renderings generated by them as LangSplat does.
Other methods focus on detail-preserving instance segmentation either by operating on the 3D map \citep{alegret2026gala} or by consolidating 2D masks from multiple images \citep{lu2025segment,nguyen2024open3dis,yan2024maskclustering,yin2026semantic}. Given semantic maps, \citet{gao2026uncertainty} support vision-language navigation, while \citet{kim2026splatreasoner} generate novel views to facilitate queries.

Online formulations include LatentBKI \citep{wilson2025latentbki} which is based on a moving average scheme to update semantics in latent space.
OpenGS-SLAM \citep{yang2025opengs-slam} and OpenGS-Fusion \citep{yang2025opengs-fusion} infer consistent labels for the objects via the multi-view consensus of 2D foundational models. 
OpenMonoGS-SLAM \citep{yoo2025openmonogs} relies on MASt3R-SLAM \citep{murai2025mast3r} for point maps, SAM for instance masks, and CLIP to generate features for each mask, which are integrated on the Gaussians.
More recently, OVI-MAP \citep{deng2026ovi} build a class-agnostic 3D instance map without using SAM.
Online Language Splatting \citep{katragadda2025_onlinelang} models both objects and background, aided by a trainable online autoencoder that compresses CLIP features to supervise language Gaussians that are disentangled from the ones modeling appearance. 

\textbf{Active Open-vocabulary Mapping.} The most relevant prior work differs from ActiveLang by only modeling objects or by resorting to frontier-based exploration without formulating a semantic uncertainty criterion.
OpenFrontier \citep{padilla2026openfrontier} is a zero-shot navigation framework that detects frontiers in the image enabling language-driven semantic tasks without dense 3D reconstruction.
RayFronts \citep{alama2025rayfronts} distinguishes between regions of the map that are within and beyond depth sensing range, relies on frontiers for exploration, but ignores background categories.
RoboAtlas \citep{schperberg2026roboatlas} uses a mixture of experts comprising frontier-based exploration, semantic mapping and egocentric VLM modules. RoboAtlas facilitates language-driven navigation but only considers object instances.

\begin{figure}[!t]
    \centering
    \includegraphics[width=0.95\textwidth]{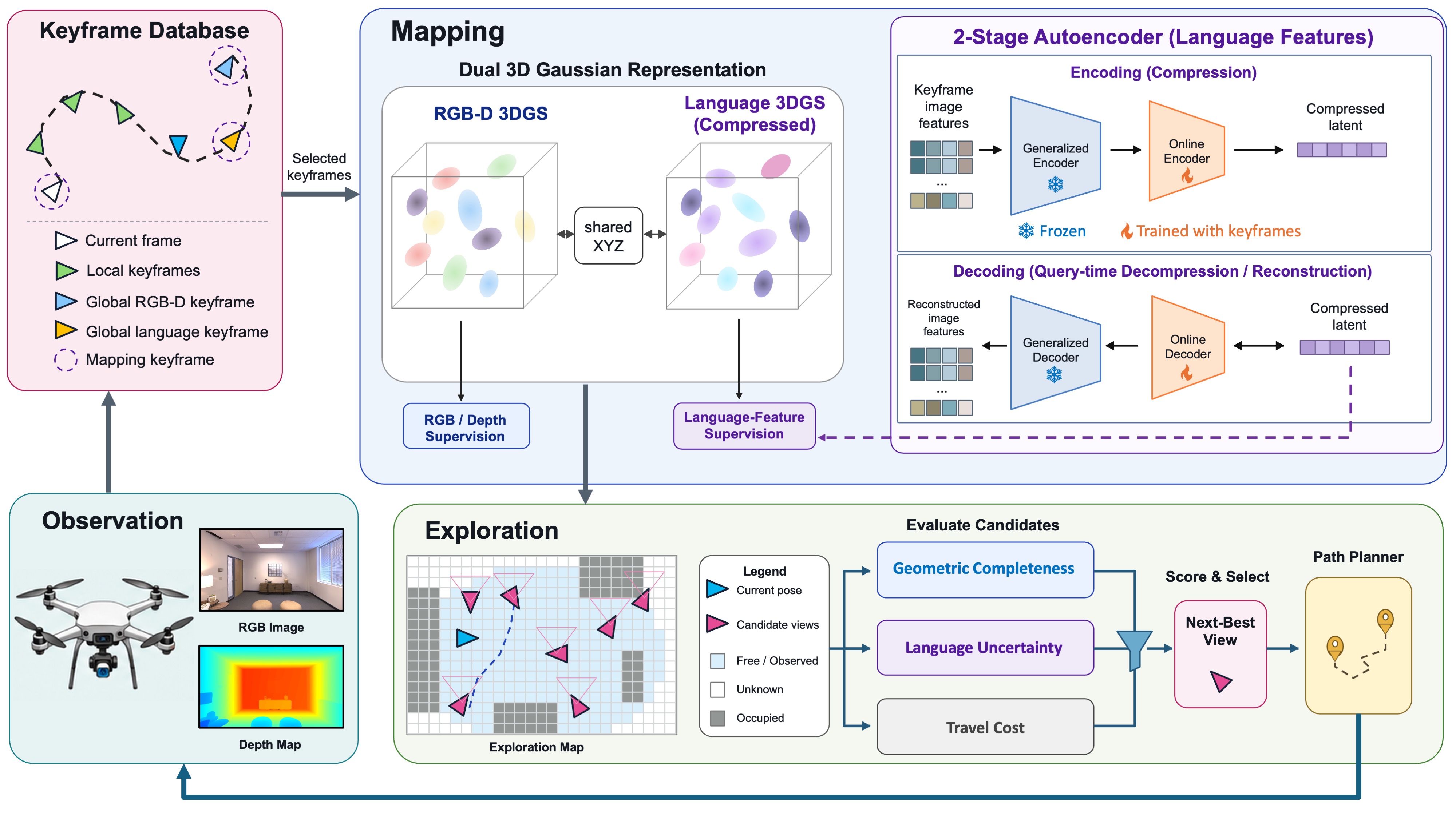}
    \vspace{-0.25cm}
    \caption{
    \textbf{Overview of the \methodname{} System.}  
    }
    \label{fig:overview}
    \vspace{-15pt}
\end{figure}

\section{Method}\label{sec:method}

\textbf{Task Definition.}
We consider active open-vocabulary 3D mapping in an unknown environment, where a robot must autonomously acquire observations to incrementally reconstruct scene geometry, appearance, and dense open-vocabulary semantics. 
Whereas passive mapping follows a predefined trajectory, our system actively determines where to observe next based on the current map. Mapping and exploration are therefore coupled: newly acquired observations update the geometric and semantic map, while the evolving map guides subsequent viewpoint selection.

\textbf{System Overview.}
The ActiveLang system is illustrated in Fig.~\ref{fig:overview}.
ActiveLang forms a closed loop between two core processes: online 3D mapping and active exploration.
Given a newly acquired RGB-D frame and its pose, the mapping module incrementally updates a dual-Gaussian representation of scene geometry and appearance in one branch, and semantic features and their uncertainty in the other.
A two-stage autoencoder, which is finetuned online, adapts the semantic representation to the sequentially observed scene.
The keyframe database retains informative RGB-D frames and provides complementary supervision for geometric, appearance, and semantic map refinement.
Using the updated map, the exploration module evaluates candidate viewpoints according to geometric coverage, semantic uncertainty, and travel cost.
The selected viewpoint produces a new observation that is incorporated into the map and closes the mapping--exploration loop.

\subsection{3DGS Representation}
\label{sec:dual_3dgs}

\textbf{Dual Gaussian Representation.}
We represent the scene using paired RGB-D and semantic Gaussian primitives that jointly encode geometry, appearance, and open-vocabulary semantics, as in Online Language Splatting~\citep{katragadda2025_onlinelang}.
For each primitive $i$, the two branches share a common 3D center $\boldsymbol{\mu}_i$ while maintaining branch-specific scale, rotation, and opacity parameters.
Specifically, following \citet{kerbl20233dgs}, the RGB-D Gaussian is parameterized as
$\mathcal{G}^{\mathrm{rgb}}_i =
\{\boldsymbol{\mu}_i,\mathbf{s}^{\mathrm{rgb}}_i,
\mathbf{q}^{\mathrm{rgb}}_i,\alpha^{\mathrm{rgb}}_i,\mathbf{c}_i\}$,
where $\mathbf{c}_i\in\mathbb{R}^3$ denotes its color.
The corresponding semantic Gaussian is represented as
$\mathcal{G}^{\mathrm{sem}}_i =
\{\boldsymbol{\mu}_i,\mathbf{s}^{\mathrm{sem}}_i,
\mathbf{q}^{\mathrm{sem}}_i,\alpha^{\mathrm{sem}}_i,
\mathbf{f}_i,u_i\}$,
where $\mathbf{f}_i\in\mathbb{R}^{15}$ is a compact semantic feature and $u_i\in\mathbb{R}$ represents semantic uncertainty.
Sharing only the Gaussian centers keeps the two representations geometrically aligned while allowing appearance and semantic information to use different spatial support.

\textbf{RGB-D Map Optimization.}
The RGB-D branch provides the geometric scaffold of the map and is optimized using incoming color and depth observations by minimizing RGB $L_1$ and DSSIM losses together with a depth $L_1$ loss over valid pixels.
The shared Gaussian centers are updated through RGB-D reconstruction, whereas semantic supervision updates only the semantic parameters.
Our RGB-D optimization follows SplaTAM~\citep{keetha2024splatam} for online Gaussian mapping.

\textbf{Semantic Features and Uncertainty.}
Each semantic Gaussian stores both a compact semantic feature and an uncertainty estimate.
The online semantic mapping module described in Sec.~\ref{sec:semantic_mapping} produces a target feature $\mathbf{f}^{\star}_p$ for each supervised pixel $p$.
Rendering the semantic Gaussians produces a semantic feature $\hat{\mathbf{f}}_p$ together with a log-variance $\hat{u}_p=\log\sigma_p^2$.
We jointly optimize semantic feature reconstruction and heteroscedastic uncertainty using
\begin{equation}
    \mathcal{L}_{\mathrm{sem}}
    = \frac{1}{|\Omega_S|}
    \sum_{p\in\Omega_S}
    \left[
        \frac{1}{2d}\exp(-\hat{u}_p)
        \left\|\mathbf{f}^{\star}_p-\hat{\mathbf{f}}_p\right\|_2^2
        + \frac{1}{2}\hat{u}_p
    \right],
    \label{eq:semantic_nll}
\end{equation}
where $\Omega_S$ denotes the supervised semantic pixels and $d=15$ is the semantic feature dimension.
The first term weighs semantic feature reconstruction according to the predicted uncertainty, while the second prevents the model from trivially increasing the uncertainty to reduce loss
\citep{kendall2017uncertainties}.
The resulting semantic uncertainty is maintained as part of the online 3D map and is subsequently used for viewpoint selection in Sec.~\ref{sec:exploration}.

\textbf{Rendering across Observation Scales.}
During exploration, the same surfaces may be viewed from substantially different distances, resulting in varying effective sampling rates. Standard 3DGS rendering may therefore exhibit scale-dependent aliasing, dilation, and erosion artifacts as the viewpoint changes.
To make both appearance and semantic rendering robust to scale variations, we apply 3D smoothing and 2D Mip filtering to both Gaussian branches.
The 3D filter is updated online as new observations are incorporated into the map as in Mip-Splatting \citep{yu2024mipsplatting}.

\subsection{Online Open-Vocabulary Semantic Mapping}
\label{sec:semantic_mapping}

Active exploration involves a temporally variable and semantically imbalanced stream of observations.
As the camera moves through different regions, large homogeneous surfaces may be repeatedly observed while less frequent semantic content causes fewer updates to the autoencoder.
This makes indiscriminate online adaptation undesirable, since the compact semantic representation can become biased toward dominant feature modes or lose distinctions between previously separable concepts.
We therefore combine online semantic feature adaptation with feature-mode-balanced compression and semantic-aware observation selection.

{\bf Online Semantic Feature Adaptation.}
To maintain a compact semantic field while adapting to newly observed content, we use an online scene-adaptive feature compression module.
For each observation, dense 768D semantic features are first compressed by a frozen general-purpose encoder into 32D codes.
A trainable online autoencoder (OLAE) further maps these codes into the 15D features stored by the semantic Gaussians.
For an intermediate feature $\mathbf{z}\in\mathbb{R}^{32}$, the online encoder and decoder produce
$\mathbf{f}=E_\theta(\mathbf{z})\in\mathbb{R}^{15}$
and
$\hat{\mathbf{z}}=D_\phi(\mathbf{f})$.
We optimize the reconstruction objective
\begin{equation}
    \ell_{\mathrm{rec}}(\mathbf{z})
    =
    \frac{1}{32}\|\hat{\mathbf{z}}-\mathbf{z}\|_1
    +
    \lambda_{\mathrm{cos}}
    \left[
        1-\operatorname{cos}(\hat{\mathbf{z}},\mathbf{z})
    \right],
    \label{eq:semantic_reconstruction}
\end{equation}
which encourages both feature fidelity and directional consistency in the frozen feature space.
We adopt the two-stage compression architecture of Online Language Splatting~\citep{katragadda2025_onlinelang}, while adapting its online training to the actively acquired observation stream as described below.

{\bf Prototype-Balanced and Separation-Preserving Adaptation.}
We address two failure modes of online semantic compression: feature-mode imbalance and compression-induced mode collapse.
To reduce feature-mode imbalance, we maintain a bounded, class-agnostic feature-mode prototype bank
$\mathcal{P}_t=\{\mathbf{b}_k\}_{k=1}^{K_t}$
in the frozen 32D feature space.
Each prototype represents a recurring semantic feature mode rather than a predefined semantic class.
Incoming features are matched to existing prototypes by cosine similarity and merged through exponential moving averages.
Unmatched feature modes are inserted into the bank, replacing the least recently used prototype when the bank is full.

At each adaptation step, we assign equal weights to the retained prototypes and combine their reconstruction loss with the expected loss of samples from the current observation:
\begin{equation}
    \mathcal{L}_{\mathrm{balanced}}
    =
    \frac{\eta}{K_t}
    \sum_{k=1}^{K_t}
    \ell_{\mathrm{rec}}(\mathbf{b}_k)
    +
    (1-\eta)
    \mathbb{E}_{\mathbf{z}\sim\mathcal{Q}_t}
    [\ell_{\mathrm{rec}}(\mathbf{z})],
    \label{eq:prototype_balanced_loss}
\end{equation}
where $\mathcal{Q}_t$ denotes the pixel sampling distribution and $\eta$ controls the balance between retained feature modes and the current samples.
Equal prototype weighting reduces the dominance of large image regions, while pixel samples preserve within-mode variation.

Balancing feature modes does not by itself prevent distinct semantics from becoming overly similar after compression.
We therefore regularize prototype pairs that are sufficiently dissimilar in the frozen feature space.
Let
$\mathcal{A}_t=
\{(i,j):i<j,\operatorname{cos}(\mathbf{b}_i,\mathbf{b}_j)<\tau_{\mathrm{in}}\}$.
For these pairs, we penalize excessive similarity between their compressed latent features:

\begin{equation}
    \mathcal{L}_{\mathrm{sep}}
    =
    \frac{1}{|\mathcal{A}_t|}
    \sum_{(i,j)\in\mathcal{A}_t}
    \left[
        \max(0, \
        \operatorname{cos}
        \left(
            E_\theta(\mathbf{b}_i),
            E_\theta(\mathbf{b}_j)
        \right)
        -m
        )
   \right]
   ,
    \label{eq:latent_separation}
\end{equation}
where 
$m$ is a similarity margin.
The OLAE is optimized using
$\mathcal{L}_{\mathrm{OLAE}}
=
\mathcal{L}_{\mathrm{balanced}}
+
\lambda_{\mathrm{sep}}\mathcal{L}_{\mathrm{sep}}$.
The encoded semantic features are detached before supervising the semantic Gaussians, separating feature adaptation from Gaussian map optimization.

{\bf Keyframe Management and Semantic Supervision Gating.}
We maintain a keyframe database with three complementary types of observations.
Local keyframes support optimization of recently observed regions, global RGB-D keyframes enable re-optimizing views with poor geometric or appearance reconstruction, and global semantic keyframes play the same role for views with large semantic reconstruction errors.
Both types of global keyframes are kept in a shared pool.

Not every actively acquired observation provides suitable supervision for online semantic adaptation.
We therefore apply OLAE and semantic-Gaussian updates only to views with sufficient feature diversity and observation context, rejecting semantically uninformative or extreme close-up views while still using them for RGB-D reconstruction.

\subsection{Semantic-Uncertainty-Guided Exploration}
\label{sec:exploration}

We sample candidate viewpoints in observed free space and score them using travel cost, geometric coverage, and semantic uncertainty; candidate-generation details are provided in the Appendix~\ref{app:impl}.
\begin{itemize}[leftmargin=*,nosep]
    \item \textbf{Travel cost.}
    $l_{\mathrm{dist}}(v)=\|\mathbf{t}_v-\mathbf{t}_t\|_2$
    measures the straight-line distance between the candidate and current camera centers.

    \item \textbf{Geometric coverage.}
    We threshold the rendered silhouette to obtain a valid-pixel mask $M_v(p)$.
    The novel-pixel count
    $n(v)=\sum_{p\in\Omega}[1-M_v(p)]$
    favors views of incompletely mapped regions, as in 
SplaTAM~\citep{keetha2024splatam}.

    \item \textbf{Semantic uncertainty.}
    $l_u(v)$ is the mean rendered semantic log-variance over valid pixels, favoring views whose semantic representation remains uncertain.
\end{itemize}

We prioritize geometric coverage before incorporating semantic uncertainty, which requires sufficient map support.
Let
$c(v)=|\Omega|^{-1}\sum_{p\in\Omega}M_v(p)$
denote view coverage.
When its mean over the candidate set is below 0.5, selection uses only travel cost and novel-pixel count.
Once sufficient map coverage is established, semantic uncertainty is additionally incorporated into viewpoint selection.
This condition is re-evaluated at every planning step.

To balance score magnitudes, we normalize the criteria across candidates:
$\tilde l_{\mathrm{dist}}=\operatorname{softmax}(l_{\mathrm{dist}})$
and
$\tilde n=\operatorname{softmax}(\log\max(n,10^{-6}))$.
We use
$\tilde l_u=\operatorname{softmax}(l_u)$
when the coverage condition is met, and
$\tilde l_u(v)=1$
otherwise.
The next view is selected as
\begin{equation}
    v^*=\argmax_{v\in\mathcal{V}_t}
    \left(1-\tilde l_{\mathrm{dist}}(v)\right)
    \tilde n(v)\,\tilde l_u(v).
    \label{eq:next_best_view}
\end{equation}
Candidates without valid pixels are assigned the minimum available semantic uncertainty.
If no candidate contains valid pixels, the semantic term is omitted.
An RRT-based planner then computes a path to the selected viewpoint.

\textbf{Observation Orientation Constraints.}
The language-aligned visual features used in our map do not explicitly account for gravity alignment or 3D geometric consistency.
Semantic predictions can therefore be sensitive to camera orientation.
Tilted observations may introduce inconsistent semantic supervision for orientation-dependent categories such as floor and ceiling.
We restrict the camera to 4-DoF motion, allowing 3D translation and yaw while fixing roll and pitch to maintain upright observations.
Our ablation study assesses the effect of this constraint.

\section{Experiments}\label{sec:exp}
\begin{table*}[t]
\vspace{-10pt}
  \centering
  \scriptsize
    \caption{\textbf{Open-vocabulary segmentation and localization in 2D and 3D on Replica and ScanNet++.}
  Results are averaged over 8 scenes for each dataset, evaluating held-out SLAM views, zero-shot novel views (NVS), and 3D point clouds, noting that \emph{Trained on same frames} highlights an inherent baseline advantage where autoencoders were fitted directly to the SLAM trajectory. Complete results are shown in Appendix Tables~\ref{tab:replica_per_scene} and~\ref{tab:scannetpp_per_scene}.}
  \vspace{5pt}
  \resizebox{0.8\linewidth}{!}{%
  \begin{tabular}{l cc @{\hspace{10pt}} cc cc @{\hspace{10pt}} cc}
    \toprule
    & \multicolumn{2}{c}{AE Settings} & \multicolumn{2}{c}{2D\,/\,SLAM}
    & \multicolumn{2}{c}{2D\,/\,NVS} & \multicolumn{2}{c}{3D} \\
    \cmidrule(lr){2-3} \cmidrule(lr){4-5} \cmidrule(lr){6-7} \cmidrule(lr){8-9}
    Method
    & \makecell{Trained on\\same frames} & Type
    & mIoU $\uparrow$ & Loc $\uparrow$ & mIoU $\uparrow$ & Loc $\uparrow$
    & mIoU $\uparrow$ & mAcc $\uparrow$ \\
    \midrule
    \multicolumn{9}{c}{\textbf{Replica}} \\
    \midrule
    LangSplat
      & \checkmark & pretrained
      & 32.36 & 63.29 & 27.65 & 61.52 & 22.89 & 35.54  \\
    OnlineLangSplat
      & \checkmark & online
      & 40.55 & 69.94 & 37.81 & 63.85 & 26.56 & 35.76  \\
    \rowcolor{gray!12}
    \methodname{}
      & \texttimes & online
      & \textbf{42.05} & \textbf{73.64} & \textbf{43.66} & \textbf{70.71}
      & \textbf{30.91} & \textbf{41.67}  \\
    \midrule
    \multicolumn{9}{c}{\textbf{ScanNet++}} \\
    \midrule
    LangSplat
      & \checkmark\ & pretrained
      & 9.06 & 24.26 & 9.06 & 23.85 & 12.22 & 23.24  \\
    OnlineLangSplat
      & \checkmark\ & online
      & 8.43 & 33.35 & 9.22 & 30.18 & 16.24 & 23.71  \\
    \rowcolor{gray!12}
    \methodname{}
      & \texttimes\ & online
      & \textbf{21.01} & \textbf{38.29} & \textbf{20.13} & \textbf{33.89}
      & \textbf{21.48} & \textbf{30.49}  \\
    \bottomrule
  \end{tabular}%
  }  
  \vspace{-4pt}
  \label{tab:semantic_openvocab}
  \label{tab:replica_openvocab}
  \label{tab:scannetpp_openvocab}
\end{table*}

\begin{table}[t]
\vspace{-10pt}
  \centering
  \scriptsize
  \caption{\textbf{Reconstruction quality on Replica.}
  We report novel-view rendering quality and 3D geometric reconstruction.
  }
  \vspace{5pt}

  \setlength{\tabcolsep}{3.5pt}
  \resizebox{0.8\linewidth}{!}{%
  \begin{tabular}{l cccc ccc}
    \toprule
    & \multicolumn{4}{c}{Novel-View Rendering}
    & \multicolumn{3}{c}{3D Geometry} \\
    \cmidrule(lr){2-5}
    \cmidrule(lr){6-8}
    Method
    & PSNR $\uparrow$
    & SSIM $\uparrow$
    & LPIPS $\downarrow$
    & L1 (cm) $\downarrow$
    & Acc (cm) $\downarrow$
    & Comp (cm) $\downarrow$
    & Comp-ratio (\%) $\uparrow$ \\
    \midrule

    LangSplat
    & 27.55 & 0.95 & 0.15 & 7.08
    & 2.48 & 2.58 & 96.35 \\

    OnlineLangSplat
    & 26.13 & 0.93 & 0.21 & 7.24
    & 2.48 & 6.37 & 80.54 \\

    \midrule

    ActiveSGM
    & 30.61 & 0.96 & 0.14 & 1.36
    & \textbf{1.19} & 1.59 & \textbf{96.68} \\

    ActiveGAMER
    & \textbf{32.02} & \textbf{0.97} & \textbf{0.11} & \textbf{1.12}
    & \textbf{1.19} & \textbf{1.56} & 96.50 \\

    \midrule

    \rowcolor{gray!12}
    \methodname{}
    & 31.77 & \textbf{0.97} & 0.13 & 1.17
    & 1.84 & 2.50 & 92.92 \\

    \bottomrule
  \end{tabular}%
  }
  \vspace{-15pt}

  \label{tab:replica_recon}
\end{table}

\textbf{Datasets and Evaluation.} 
We use Habitat~\citep{savva2019habitat} for RGB-D rendering and a dense OpenCLIP extractor~\citep{cherti2023reproducible} to obtain pixel-aligned semantic features.
We evaluate on eight scenes from Replica~\citep{straub2019replica} and eight scenes from ScanNet++~\citep{yeshwanth2023scannetpp}.
For both datasets, all methods are evaluated on held-out views from the predefined capture trajectory (\emph{SLAM}), zero-shot novel views (\emph{NVS}), and directly on the reconstructed 3D map.

For open-vocabulary semantic mapping, we use the ten most frequent semantic categories in each scene as text queries.
We report 2D mean Intersection over Union (mIoU) and localization accuracy (Loc), together with 3D mIoU and mean class accuracy (mAcc).
For reconstruction quality, we report geometric accuracy, completeness, and completeness ratio, as well as PSNR, SSIM, LPIPS, and depth $L_1$ for novel-view rendering.
Details on all aspects are provided in the Appendix.

\textbf{Baselines.}
We compare primarily with LangSplat~\citep{qin2024langsplat} and Online Language Splatting~\citep{katragadda2025_onlinelang} as representative offline and online, passive, open-vocabulary mapping baselines.\textit{ Note that there are no active open-vocabulary mapping baselines to compare against.}
For geometric reconstruction and novel-view synthesis, we additionally compare with ActiveSGM~\citep{chen2025understanding} and ActiveGAMER~\citep{chen2025activegamer}.

\begin{figure}[tb]
    \centering
    \includegraphics[width=0.95\textwidth]{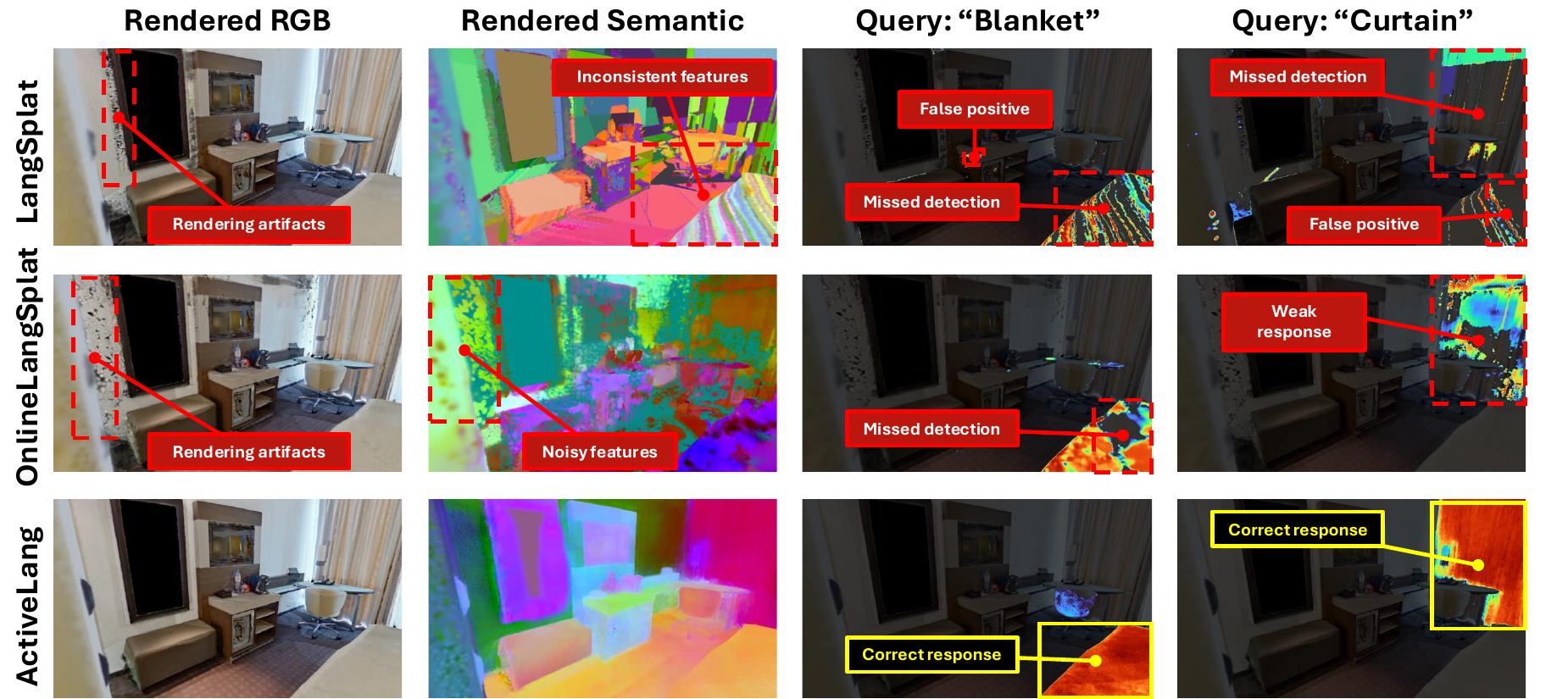}
    \vspace{-0.25cm}
    \caption{\textbf{Qualitative results for ScanNet++ on scene 6115eddb86.}
Rows show LangSplat, OnlineLangSplat, and \methodname{} (ours), from top to bottom.
Columns show rendered RGB, semantic maps, and responses to the language queries ``blanket'' and ``curtain.''
Red dashed boxes highlight rendering artifacts, semantic errors, and missed, spurious, or weak query responses; yellow boxes indicate correct responses. Our method generates more complete queried object.}
    \label{fig:scannetpp_qual}
    \vspace{-15pt}
\end{figure}

\subsection{Active Open-Vocabulary Mapping}
\label{sec:active_openvocab_mapping}

Table~\ref{tab:semantic_openvocab} compares \methodname{} with LangSplat and OnlineLangSplat on Replica and ScanNet++ across held-out SLAM views, zero-shot novel views, and the reconstructed 3D maps.

\textbf{2D Open-Vocabulary Mapping.}
Across both datasets, \methodname{} consistently improves open-vocabulary segmentation and localization over the passive mapping baselines.
On Replica, the advantage becomes particularly clear on strictly novel views, where \methodname{} surpasses OnlineLangSplat by $+5.9\%$ in mIoU, indicating stronger generalization beyond the observations used during mapping.
The gain is substantially larger on ScanNet++, where \methodname{} improves 2D mIoU by a wide margin over both baselines, showing that the proposed mapping and exploration strategy remains effective under more complex real-world scene geometry and viewpoint variation.
Qualitative comparisons in Figs~\ref{fig:scannetpp_qual} and~\ref{fig:replica_semantic} further show cleaner and more spatially coherent semantic fields, with fewer missed detections or spurious query responses.

\textbf{3D Open-Vocabulary Mapping.}
The same overall trend holds when semantic quality is evaluated directly on the reconstructed 3D maps.
\methodname{} achieves the highest 3D mIoU across both datasets and improves over OnlineLangSplat in semantic accuracy, demonstrating that the gains are not limited to rendered 2D views.
The improvement is generally smaller in 3D than in 2D, suggesting that part of the advantage also comes from the quality of the rendered semantic field rather than only from per-Gaussian semantic assignments.
Overall, the results in Table~\ref{tab:semantic_openvocab} show consistent gains across datasets, viewpoints, and both 2D and 3D evaluation settings.

\subsection{Reconstruction Quality on Replica}
\label{sec:results_replica_reconstruction}

We evaluate reconstruction quality primarily on Replica, whose high-quality geometry and controlled rendering setup provide reliable ground truth for both novel-view synthesis and 3D reconstruction.

\textbf{Novel-View Synthesis.}
As shown in Table~\ref{tab:replica_recon}, \methodname{} remains competitive with more geometry-focused mapping systems while substantially outperforming the open-vocabulary baselines.
Compared with LangSplat and OnlineLangSplat, \methodname{} improves PSNR by more than $4$ dB while reducing depth error by approximately $6$ cm.
Compared with geometry-focused methods, \methodname{} surpasses ActiveSGM and performs comparably to ActiveGAMER.
These results show that incorporating dense open-vocabulary semantics and semantic-aware exploration does not substantially compromise appearance or depth reconstruction quality.

\textbf{3D Geometric Reconstruction.}
As shown in Table~\ref{tab:replica_recon}, \methodname{} achieves the best geometric accuracy among the open-vocabulary mapping baselines and substantially improves completeness over OnlineLangSplat.
Geometry-focused active mappers achieve higher completeness, partly because they can exploit unconstrained 6-DoF viewpoints to inspect non-horizontal surfaces.
In contrast, \methodname{} restricts camera motion to 4 DoF to maintain more consistent semantic observations.
Ablating this constraint to 6 DoF yields modest geometric gains but substantially degrades open-vocabulary segmentation by $8.6\%$ mIoU in 2D and $5.0\%$ mIoU in 3D.

Additional novel-view synthesis results on ScanNet++ are reported in Table~\ref{tab:scannetpp_per_scene} in the Appendix, where \methodname{} also remains competitive.

\subsection{Ablation Study}
\label{sec:ablation}
\begin{table}[t]
\vspace{-12pt}
 \caption{\textbf{Component ablation on Replica.} Averaged across four scenes with the best score in \textbf{bold} and the second best \underline{underlined}. The results show that while global keyframe variants offer minor metric trade-offs, ablating language supervision or the 4-DoF orientation constraint leads to performance drops.}
  \label{tab:ablation}
  \centering
  \setlength{\tabcolsep}{2.5pt}
  \resizebox{\linewidth}{!}{%
  \begin{tabular}{l ccc @{\hspace{6pt}} cc @{\hspace{6pt}} c @{\hspace{8pt}} cccc @{\hspace{8pt}} cc @{\hspace{8pt}} cc}
    \toprule
    & \multicolumn{3}{c}{Keyframes} & \multicolumn{2}{c}{Lang.\ supervision} & Cam.
    & \multicolumn{4}{c}{Reconstruction} & \multicolumn{2}{c}{2D\,/\,NVS} & \multicolumn{2}{c}{3D} \\
    \cmidrule(lr){2-4} \cmidrule(lr){5-6} \cmidrule(lr){7-7}
    \cmidrule(lr){8-11} \cmidrule(lr){12-13} \cmidrule(lr){14-15}
    Variant & Local & RGB-D & Lang.\ & Gate & Proto.\ & DoF
    & PSNR $\uparrow$ & SSIM $\uparrow$ & LPIPS $\downarrow$ & L1 (cm) $\downarrow$
    & mIoU $\uparrow$ & Loc $\uparrow$
    & mIoU $\uparrow$ & mAcc $\uparrow$ \\
    \midrule
    w/o prototype bank
      & \checkmark & \checkmark & \checkmark & \checkmark & \texttimes & 4
      & \textbf{33.16} & \textbf{0.97} & \textbf{0.13} & \underline{0.86} & 38.44 & 65.05 & 21.12 & 30.13 \\
    w/o supervision gate
      & \checkmark & \checkmark & \checkmark & \texttimes & \checkmark & 4
      & 32.86 & \underline{0.96} & \underline{0.14} & 0.94 & 38.13 & 64.18 & 21.13 & 29.74 \\
    \midrule
    w/o orientation constraint
      & \checkmark & \checkmark & \checkmark & \checkmark & \checkmark & 6
      & 30.71 & 0.95 & 0.16 & 1.19 & 32.33 & 61.34 & 22.39 & 36.08 \\
    \midrule
    w/o semantic uncertainty
      & \checkmark & \checkmark & \checkmark & \checkmark & \checkmark & 4
      & 32.97 & \textbf{0.97} & \textbf{0.13} & 0.89 & 39.88 & 67.11 & 27.09 & \textbf{39.27} \\
    \midrule
    local keyframes only
      & \checkmark & \texttimes & \texttimes & \checkmark & \checkmark & 4
      & 32.19 & \underline{0.96} & 0.15 & 1.18 & \textbf{41.04} & \textbf{70.01} & \textbf{27.46} & \underline{38.93} \\
    \midrule
    \rowcolor{gray!12}
    \methodname{} (full)
      & \checkmark & \checkmark & \checkmark & \checkmark & \checkmark & 4
      & \underline{33.07} & \textbf{0.97} & \textbf{0.13} & \textbf{0.85} & \underline{40.91} & \underline{67.74} & \underline{27.34} & 38.81 \\
    \bottomrule
  \end{tabular}%
  }
\vspace{-15pt}
\end{table}

Table~\ref{tab:ablation} shows our ablations of the main components of \methodname{} across four Replica scenes.

\textbf{Online Semantic Adaptation.}
Removing the prototype-based adaptation mechanism reduces 2D novel-view mIoU from $40.91\%$ to $38.44\%$ and 3D mIoU from $27.34\%$ to $21.12\%$.
Removing semantic supervision gating causes similar reductions in 2D and 3D mIoU. These results show that balancing persistent feature modes and filtering unsuitable semantic observations are important for maintaining a reliable representation during online adaptation.
Since the separation objective operates on the prototype bank, the prototype ablation evaluates the prototype-based adaptation mechanism as a whole without isolating the separation term.

\textbf{Semantic-Aware Exploration.}
Removing semantic uncertainty from viewpoint selection causes a modest decrease in semantic performance, with 2D mIoU dropping from $40.91\%$ to $39.88\%$ and 3D mIoU from $27.34\%$ to $27.09\%$.
This indicates that semantic uncertainty provides complementary guidance beyond geometric coverage when deciding where to observe.
A substantially larger degradation occurs when the 4-DoF orientation constraint is removed: 2D mIoU decreases by $8.6\%$ and PSNR drops by approximately $2.4$\,dB.
This result highlights the importance of controlling how the scene is observed, since upright viewpoints provide more consistent semantic supervision.

\textbf{Keyframe Management.}
Using only local keyframes primarily affects geometric and appearance reconstruction rather than semantic accuracy.
Depth error increases from $0.85$ to $1.18$\,cm and PSNR decreases from $33.07$ to $32.19$\,dB, while semantic metrics remain nearly unchanged.
This suggests that global keyframes mainly support continual geometric and appearance refinement. 

\section{Conclusion}\label{sec:conclusion}
We have presented \methodname{}, an active open-vocabulary 3D mapping system combining a dual-Gaussian structure, prototype-balanced online language adaptation, and semantic-uncertainty-guided exploration. 
Evaluated on Replica and ScanNet++, \methodname{} outperforms existing online and offline \textit{passive} mapping baselines in 2D and 3D open-vocabulary segmentation while matching geometry-focused mappers in visual quality at lower computational cost. 
Current limitations include assuming ground-truth poses are available and minor surface detail trade-offs imposed by camera orientation constraints. 
Future work will focus on deploying with unknown camera poses and closing the geometric gap while preserving multi-modal stability.

\clearpage

\bibliographystyle{iclr2027_conference}
\bibliography{bib/conference_name,bib/nerf,bib/semantic,bib/ActiveMapping}

\newpage

\appendix
\section{Appendix}

\setcounter{table}{0}
\setcounter{figure}{0}
\renewcommand{\thetable}{A\arabic{table}}
\renewcommand\thefigure{A\arabic{figure}}
\renewcommand{\theHtable}{A.Tab.\arabic{table}}
\renewcommand{\theHfigure}{A.Abb.\arabic{figure}}
\renewcommand\theequation{A\arabic{equation}}
\renewcommand{\theHequation}{A.Abb.\arabic{equation}}

This appendix provides additional experimental results, analyses, and implementation details. Sec.~\ref{app:results} supplements the experiments in the main paper with additional results. Secs.~\ref{app:embedding_alignment} and~\ref{app:supp_purity} further investigate image–text feature misalignment and feature-mode imbalance with compression-induced mode collapse, respectively, providing empirical evidence for the two key challenges in active open-vocabulary mapping discussed in the introduction. Sec.~\ref{app:reprod} describes the datasets, hardware, and open-source software used to reproduce our experiments. Finally, Sec.~\ref{app:impl} provides further implementation details, including the hyperparameters for each module.

\subsection{Additional Results}\label{app:results}
We present additional results for both datasets in this section, including exploration trajectories of \methodname{}, per-scene rendering quality on novel views, and 2D and 3D open-vocabulary semantic evaluation. For each scene, we select the ten most frequent semantic categories in the ground-truth mesh as text queries and use the same query text for both 2D and 3D semantic evaluation.

\begin{wrapfigure}{r}{0.45\textwidth}
    \centering
    \vspace{-14pt}
    \setlength{\parskip}{0pt}
    \setlength{\abovecaptionskip}{2pt}
    \setlength{\belowcaptionskip}{0pt}
    \includegraphics[
        width=\linewidth,
        trim=0bp 7bp 0bp 3bp,
        clip
    ]{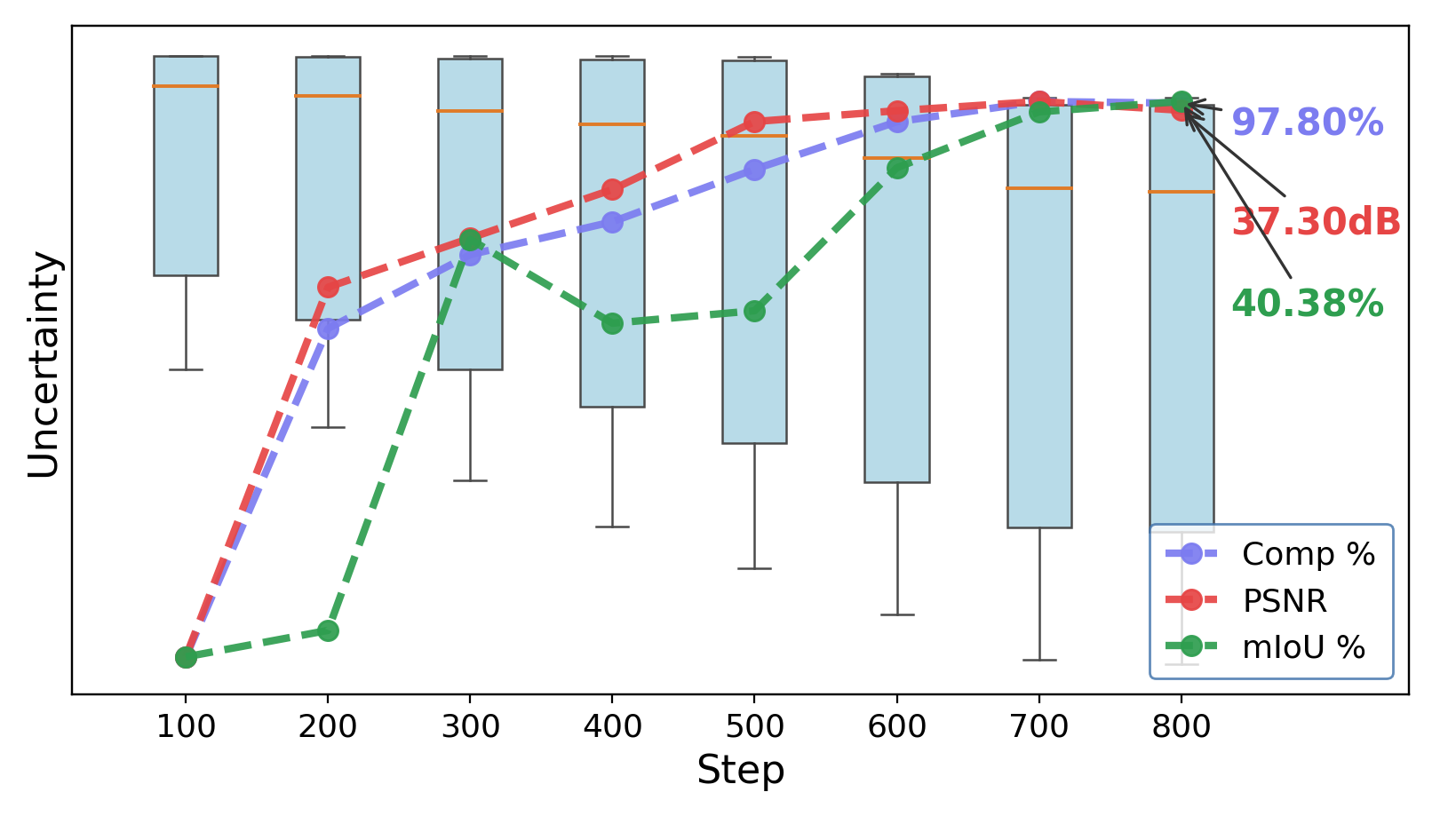}
    \vspace{-10pt}
    \caption{\textbf{Exploration progress on Replica \texttt{office1}.} Boxplots show mean rendered language uncertainty; 
    curves show PSNR, reconstruction completeness, and 2D mIoU. 
    Uncertainty decays shows an overall downward trend as rendering, geometry, and semantics improve during exploration.}
    \label{fig:of1_progress}
    \vspace{0pt}
\end{wrapfigure}

\begin{figure}[tb]
    \centering
    \begin{tabular}{c}
    \end{tabular}
    \includegraphics[width=0.9\textwidth]{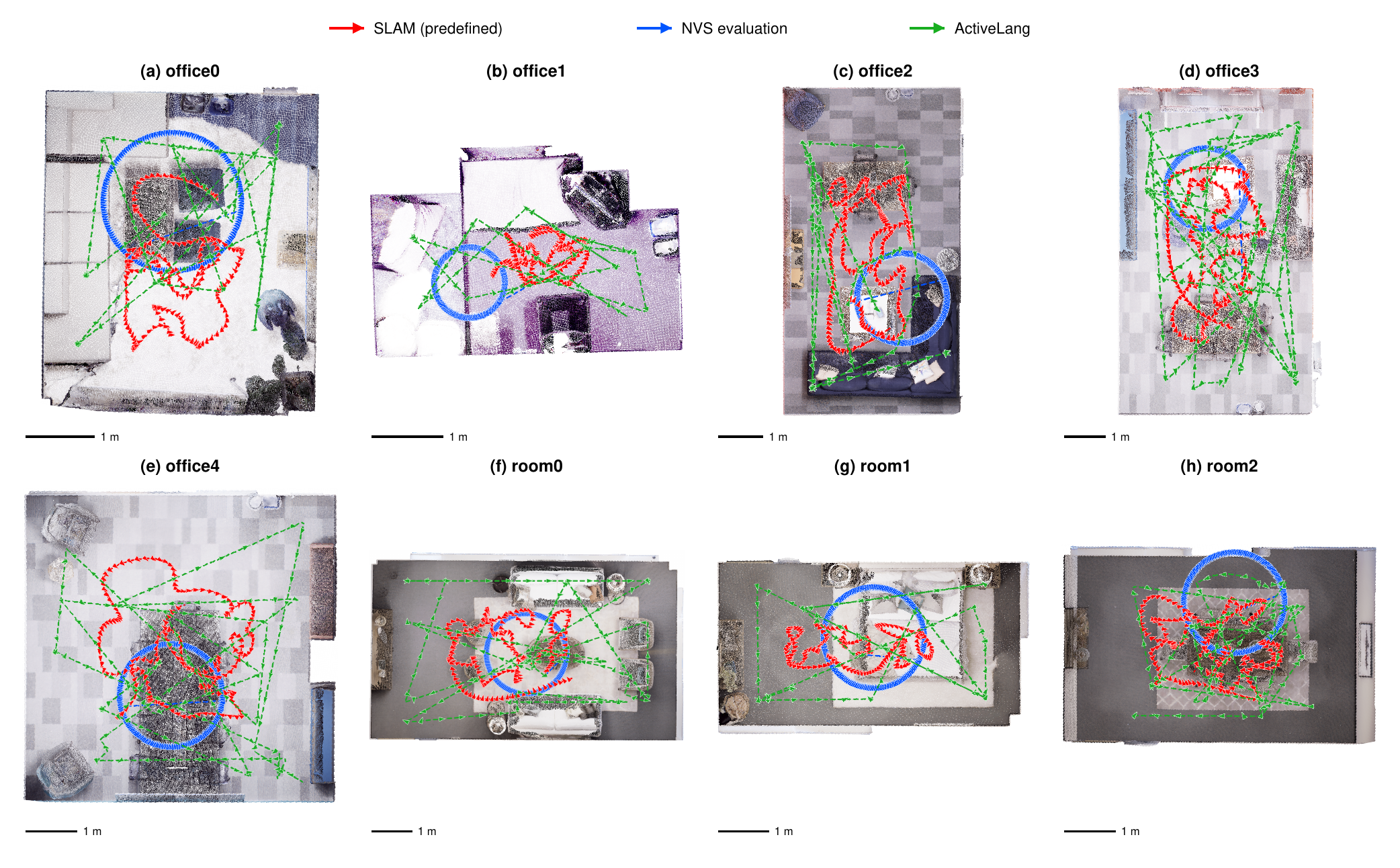} \\
    \textbf{(a) Trajectory comparison in the eight Replica scenes} \\ 
    \includegraphics[width=0.9\textwidth]{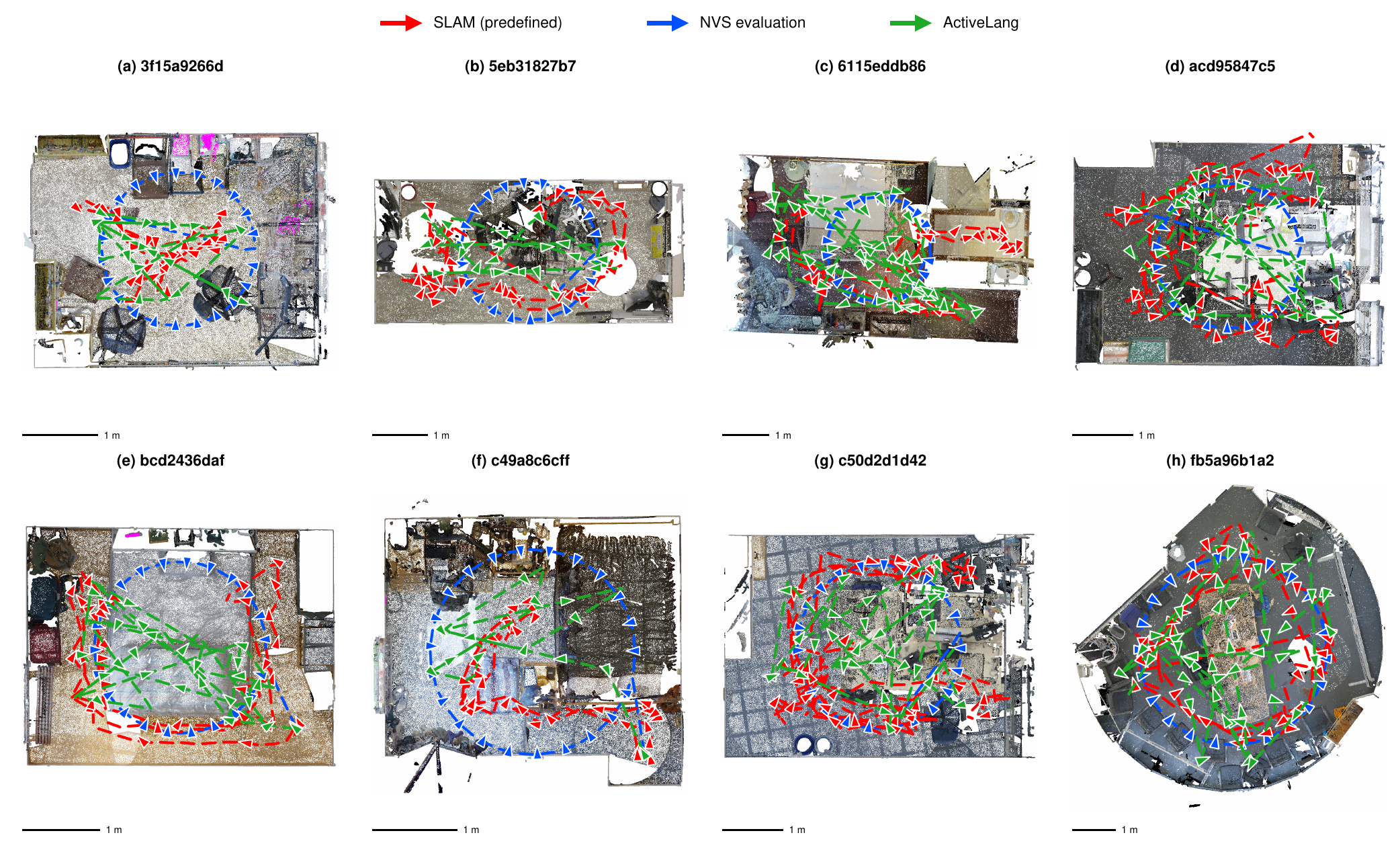} \\
    \textbf{(b) Trajectory comparison in the eight Scannet++ scenes} \\ 
    \caption{
    Red, blue, and green denote the predefined SLAM trajectory,
    the novel-view evaluation trajectory, and the trajectory generated
    by \methodname{}, respectively.
    Our active trajectories extend beyond the regions covered by the
    predefined trajectories, exploring a broader range of viewpoints
    throughout each room. Each panel includes a 1\,m scale bar.}
    \label{fig:trajectories}
    \vspace{-20pt}
\end{figure}

\paragraph{Replica.} The exploration trajectories of \methodname{} are shown as green lines in Fig.~\ref{fig:trajectories}(a), and the text queries are listed in Table~\ref{tab:replica_queries}. Table~\ref{tab:replica_per_scene} reports per-scene results for open-vocabulary semantic mapping and novel-view rendering. Our method outperforms the baselines in most scenes, particularly in the 2D semantic metrics.

We analyze how map quality and predicted language uncertainty evolve
during exploration in Replica \texttt{office1}
(Fig.~\ref{fig:of1_progress}).
We save a checkpoint every 100 exploration steps and evaluate reconstruction progress using a fixed set of 20 views uniformly sampled along the SLAM trajectory. At each checkpoint, we render RGB images, semantic maps, and
language uncertainty maps at these viewpoints.
We evaluate RGB reconstruction using PSNR and open-vocabulary
segmentation using 2D mIoU with the LERF evaluation protocol.
The boxplots summarize the distribution of mean rendered language
uncertainty across the evaluation views.
To additionally measure geometric coverage, we compute the fraction
of points sampled from the reconstructed map that lie within
5\,cm of the ground-truth mesh.
The three performance curves are independently normalized
for visualization; endpoint annotations report their original values.

\begin{table}[b]
   \vspace{-20pt}
    \centering
    \small
    \setlength{\tabcolsep}{6pt}
    \renewcommand{\arraystretch}{1.2}
    \caption{\textbf{Scene-specific text queries for Replica.}
    We use the following 10 category labels in each scene for 2D and 3D
    open-vocabulary semantic evaluation.}
    \vspace{5pt}
    \label{tab:replica_queries}
    \resizebox{0.9\textwidth}{!}{
    \begin{tabular}{@{}l p{0.83\linewidth}@{}}
        \toprule
        Scene & Query categories \\
        \midrule
        \texttt{office0} & wall, ceiling, rug, sofa, blinds, table, chair, indoor-plant, tv-screen, door \\
        \texttt{office1} & wall, ceiling, floor, pillow, blinds, pillar, desk, door, chair, blanket \\
        \texttt{office2} & wall, floor, ceiling, sofa, table, chair, lamp, tv-screen, door, stool \\
        \texttt{office3} & wall, floor, ceiling, chair, sofa, table, window, tv-stand, blinds, lamp \\
        \texttt{office4} & wall, floor, ceiling, chair, window, bench, lamp, tv-screen, door, bin \\
        \midrule
        \texttt{room0} & wall, ceiling, floor, blinds, sofa, rug, chair, table, lamp, window \\
        \texttt{room1} & wall, ceiling, floor, blinds, rug, lamp, bed, pillow, comforter, nightstand \\
        \texttt{room2} & wall, ceiling, floor, chair, blinds, rug, table, lamp, shelf, door \\
        \bottomrule
    \end{tabular}
    }
\end{table}

As exploration proceeds, PSNR and geometric completeness increase
and gradually plateau, reaching 37.30\,dB and 97.80\%, respectively.
Semantic quality improves overall to 40.38\% mIoU, although it
temporarily decreases at intermediate checkpoints, indicating that
semantic refinement is less uniform than geometric coverage.
Meanwhile, the median rendered language uncertainty decreases,
and the lower quartile shifts downward substantially.
The distribution remains broad, suggesting that uncertainty is
resolved unevenly across the evaluation views.
This overall reduction is desirable: as the agent acquires additional
informative observations, the map should become better constrained
and its semantic predictions less uncertain.
The concurrent improvements in reconstruction and segmentation
are consistent with this interpretation.

\begin{figure}[tb]
    \centering
    \includegraphics[width=0.9\textwidth]{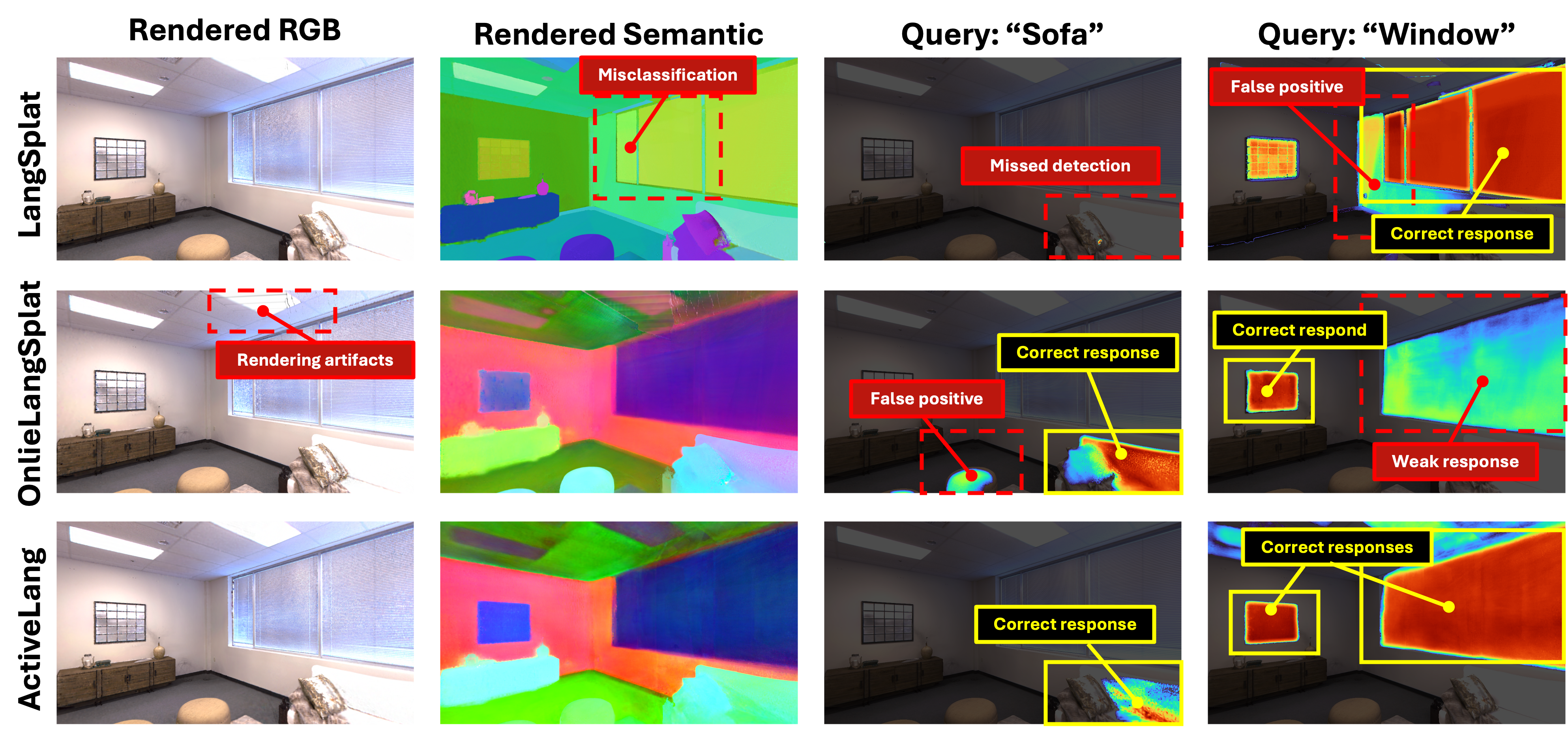}
    \vspace{-0.25cm}
    \caption{\textbf{Qualitative comparison on a view from Replica room0.}
Rows show LangSplat, OnlineLangSplat, and \methodname{} (ours), from top to bottom.
Columns show rendered RGB, semantic maps, and responses to the language queries ``sofa'' and ``window.''
Red dashed boxes highlight rendering artifacts, semantic errors, and missed, spurious, or weak query responses; yellow boxes indicate correct responses.
In the highlighted regions, our method reduces spurious responses to ``sofa'' and responds consistently to both regions queried as ``window.''}
\label{fig:replica_semantic}
    \vspace{-16pt}
\end{figure}

Figure~\ref{fig:replica_semantic} presents a qualitative comparison with LangSplat and OnlineLangSplat from a viewpoint in Replica \texttt{room0}, showing rendered RGB images, PCA visualizations of rendered language features, and relevancy maps for the queries ``sofa'' and ``window.'' In the highlighted regions, our method reduces spurious responses to the ``sofa'' query and produces more consistent responses across the ``window'' regions. For the PCA visualizations, we project the 3D features of LangSplat and the 32D features of OnlineLangSplat and our method onto their respective three principal components. PCA is fitted independently for each method, so the resulting colors are not directly comparable.

\begin{table}[t]
\vspace{-15pt}
  \centering
  \caption{\textbf{Per-scene open-vocabulary semantic mapping and rendering quality on Replica.}
  Semantic metrics are evaluated on held-out SLAM views, zero-shot novel views (NVS), and 3D point clouds
  using the ten scene-specific query categories in Table~\ref{tab:replica_queries}.
  Rendering metrics (PSNR, SSIM, and LPIPS) are evaluated on novel views (NVS).
  Results provide per-scene breakdowns of Tables~\ref{tab:replica_openvocab} and~\ref{tab:replica_recon}.
  Best result per scene and metric in \textbf{bold.}}
  \vspace{5pt}
  \scriptsize
  \setlength{\tabcolsep}{4pt}
  \resizebox{\linewidth}{!}{%
  \begin{tabular}{l l @{\hspace{10pt}} cc cc @{\hspace{10pt}} cc @{\hspace{10pt}} ccc}
    \toprule
    & & \multicolumn{2}{c}{2D\,/\,SLAM}
    & \multicolumn{2}{c}{2D\,/\,NVS} & \multicolumn{2}{c}{3D} & \multicolumn{3}{c}{Rendering} \\
    \cmidrule(lr){3-4} \cmidrule(lr){5-6} \cmidrule(lr){7-8} \cmidrule(lr){9-11}
    Scene & Method
    & mIoU $\uparrow$ & Loc $\uparrow$ & mIoU $\uparrow$ & Loc $\uparrow$
    & mIoU $\uparrow$ & mAcc $\uparrow$ & PSNR $\uparrow$ & SSIM $\uparrow$ & LPIPS $\downarrow$ \\
    \midrule
    \multirow{3}{*}{\texttt{office0}}
      & LangSplat       & 30.97 & 63.37 & 27.33 & 52.83 & \textbf{27.03} & \textbf{40.38} & 29.96 & 0.952 & 0.123 \\
      & OnlineLangSplat & \textbf{42.43} & \textbf{73.80} & \textbf{36.03} & \textbf{53.66} & 23.04 & 26.12 & 27.54 & 0.921 & 0.228 \\
    \rowcolor{gray!12}
      & \methodname{}   & 32.78 & 71.77 & 30.08 & 52.60 & 19.43 & 26.37 & \textbf{35.13} & \textbf{0.979} & \textbf{0.092} \\
    \midrule
    \multirow{3}{*}{\texttt{office1}}
      & LangSplat       & \textbf{30.14} & \textbf{64.96} & 25.68 & 59.90 & 22.65 & 35.39 & 31.34 & 0.963 & 0.137 \\
      & OnlineLangSplat & 21.47 & 48.09 & 32.06 & \textbf{62.48} & 12.62 & 22.76 & 29.09 & 0.934 & 0.251 \\
    \rowcolor{gray!12}
      & \methodname{}   & 27.00 & 60.63 & \textbf{39.14} & 61.93 & \textbf{25.73} & \textbf{45.92} & \textbf{37.23} & \textbf{0.978} & \textbf{0.115} \\
    \midrule
    \multirow{3}{*}{\texttt{office2}}
      & LangSplat       & 38.87 & \textbf{75.31} & 30.07 & \textbf{76.72} & 21.79 & \textbf{33.47} & 23.92 & 0.930 & 0.194 \\
      & OnlineLangSplat & 44.27 & 71.28 & 31.31 & 50.79 & 22.54 & 28.41 & 22.26 & 0.897 & 0.288 \\
    \rowcolor{gray!12}
      & \methodname{}   & \textbf{49.99} & 71.95 & \textbf{45.96} & 72.72 & \textbf{30.17} & 32.72 & \textbf{29.10} & \textbf{0.962} & \textbf{0.154} \\
    \midrule
    \multirow{3}{*}{\texttt{office3}}
      & LangSplat       & 29.43 & 63.36 & 22.25 & 48.04 & 13.53 & 20.35 & 25.87 & 0.932 & 0.224 \\
      & OnlineLangSplat & 50.00 & 85.47 & 51.03 & \textbf{92.12} & \textbf{35.62} & \textbf{39.99} & 23.61 & 0.889 & 0.314 \\
    \rowcolor{gray!12}
      & \methodname{}   & \textbf{51.41} & \textbf{86.95} & \textbf{52.83} & 80.04 & 32.04 & 39.85 & \textbf{30.46} & \textbf{0.973} & \textbf{0.134} \\
    \midrule
    \multirow{3}{*}{\texttt{office4}}
      & LangSplat       & 28.30 & 58.42 & 19.93 & 58.60 & 16.70 & 32.95 & 30.31 & 0.975 & \textbf{0.101} \\
      & OnlineLangSplat & 40.14 & 67.54 & 35.51 & 51.24 & 24.59 & 42.76 & 28.48 & 0.961 & 0.187 \\
    \rowcolor{gray!12}
      & \methodname{}   & \textbf{44.71} & \textbf{75.67} & \textbf{44.78} & \textbf{74.36} & \textbf{34.32} & \textbf{50.91} & \textbf{32.76} & \textbf{0.981} & 0.103 \\
    \midrule
    \multirow{3}{*}{\texttt{room0}}
      & LangSplat       & 32.45 & 59.13 & 34.86 & 64.25 & 30.99 & 42.88 & \textbf{30.01} & \textbf{0.960} & \textbf{0.123} \\
      & OnlineLangSplat & 44.99 & 73.46 & 44.50 & 79.92 & 36.22 & 51.08 & 28.76 & 0.947 & 0.167 \\
    \rowcolor{gray!12}
      & \methodname{}   & \textbf{47.43} & \textbf{79.53} & \textbf{51.47} & \textbf{82.99} & \textbf{36.95} & \textbf{53.29} & 29.60 & 0.953 & 0.162 \\
    \midrule
    \multirow{3}{*}{\texttt{room1}}
      & LangSplat       & 35.75 & 60.07 & 34.91 & 63.98 & 24.57 & \textbf{36.67} & 28.18 & \textbf{0.957} & \textbf{0.112} \\
      & OnlineLangSplat & 34.81 & 59.13 & 37.03 & 60.62 & 24.24 & 31.24 & 28.18 & 0.955 & 0.114 \\
    \rowcolor{gray!12}
      & \methodname{}   & \textbf{41.80} & \textbf{70.43} & \textbf{42.96} & \textbf{73.45} & \textbf{27.24} & 29.67 & \textbf{30.31} & 0.956 & 0.139 \\
    \midrule
    \multirow{3}{*}{\texttt{room2}}
      & LangSplat       & 32.94 & 61.71 & 26.19 & \textbf{67.83} & 25.87 & 42.22 & 20.81 & 0.921 & 0.162 \\
      & OnlineLangSplat & 39.66 & 59.21 & 34.98 & 59.97 & 33.61 & 43.76 & 21.09 & 0.921 & 0.162 \\
    \rowcolor{gray!12}
      & \methodname{}   & \textbf{41.26} & \textbf{72.22} & \textbf{42.05} & 67.60 & \textbf{41.38} & \textbf{54.62} & \textbf{29.55} & \textbf{0.953} & \textbf{0.134} \\
    \bottomrule
  \end{tabular}%
  }
  \vspace{-20pt}
  \label{tab:replica_per_scene}
  \label{tab:replica_per_scene_recon}
\end{table}

\begin{table}[b]
\vspace{-15pt}
    \centering
    \small
    \setlength{\tabcolsep}{6pt}
    \renewcommand{\arraystretch}{1.2}
    \caption{\textbf{Scene-specific text queries for ScanNet++.}
    We use the following 10 category labels in each scene for
    open-vocabulary semantic evaluation.}
    \label{tab:scannetpp_queries}
    \vspace{5pt}
    \resizebox{0.9\textwidth}{!}{
    \begin{tabular}{@{}l p{0.86\linewidth}@{}}
        \toprule
        Scene & Query categories \\
        \midrule
        \texttt{3f15a9266d} & wall, blinds, floor, ceiling, cabinet, bookshelf, jacket, ceiling lamp, window, monitor \\
        \texttt{5eb31827b7} & wall, ceiling, floor, monitor, door, window, whiteboard, table, ceiling lamp, office chair \\
        \texttt{6115eddb86} & curtain, wall, floor, bed, ceiling, shower wall, ceiling lamp, door, blanket, plant \\
        \texttt{acd95847c5} & floor, cabinet, ceiling, wall, ceiling lamp, shelf, bookshelf, table, pipe, window \\
        \texttt{bcd2436daf} & wall, ceiling, shower wall, floor, blanket, window, heater, ceiling lamp, bookshelf, tv \\
        \texttt{c49a8c6cff} & wall, floor, shoes, chair, blinds, monitor, door, curtain, table, ceiling \\
        \texttt{c50d2d1d42} & wall, floor, ceiling, ceiling lamp, refrigerator, shelf, chair, monitor, blinds, whiteboard \\
        \texttt{fb5a96b1a2} & ceiling, floor, wall, bookshelf, whiteboard, table, cabinet, doorframe, ceiling lamp, window frame \\
        \bottomrule
    \end{tabular}
    }
\end{table}

\begin{table}[tb]
  \centering
    \caption{\textbf{Per-scene open-vocabulary semantic mapping and rendering quality on ScanNet++.}
  Semantic results provide a per-scene breakdown of Table~\ref{tab:scannetpp_openvocab}.
  2D metrics are computed on 50 held-out SLAM views and a 180-view novel-view orbit (NVS),
  and 3D metrics against the ground-truth semantic mesh, using the ten scene-specific query categories
  in Table~\ref{tab:scannetpp_queries}.
  Rendering metrics (PSNR, SSIM, and LPIPS) are evaluated on the 180-view novel-view orbit (NVS).
  Best result per scene and metric in \textbf{bold}.}
  \vspace{5pt}
  \scriptsize
  \setlength{\tabcolsep}{4pt}
  \resizebox{\linewidth}{!}{%
  \begin{tabular}{l l @{\hspace{10pt}} cc cc @{\hspace{10pt}} cc @{\hspace{10pt}} ccc}
    \toprule
    & & \multicolumn{2}{c}{2D\,/\,SLAM}
    & \multicolumn{2}{c}{2D\,/\,NVS} & \multicolumn{2}{c}{3D} & \multicolumn{3}{c}{Rendering} \\
    \cmidrule(lr){3-4} \cmidrule(lr){5-6} \cmidrule(lr){7-8} \cmidrule(lr){9-11}
    Scene & Method
    & mIoU $\uparrow$ & Loc $\uparrow$ & mIoU $\uparrow$ & Loc $\uparrow$
    & mIoU $\uparrow$ & mAcc $\uparrow$ & PSNR $\uparrow$ & SSIM $\uparrow$ & LPIPS $\downarrow$ \\
    \midrule
    \multirow{3}{*}{\texttt{3f15a9266d}}
      & LangSplat       & 10.25 & 25.82 & 11.36 & 24.74 & 9.85 & 18.61 & 26.23 & \textbf{0.961} & \textbf{0.109} \\
      & OnlineLangSplat & 16.38 & 48.68 & 15.31 & 49.14 & 19.07 & 25.86 & 24.30 & 0.921 & 0.209 \\
    \rowcolor{gray!12}
      & \methodname{}   & \textbf{33.47} & \textbf{54.03} & \textbf{35.78} & \textbf{53.66} & \textbf{24.80} & \textbf{36.55} & \textbf{26.57} & 0.954 & 0.121 \\
    \midrule
    \multirow{3}{*}{\texttt{5eb31827b7}}
      & LangSplat       & 16.28 & 25.68 & 10.09 & 28.20 & 10.93 & 24.29 & 17.47 & 0.782 & 0.355 \\
      & OnlineLangSplat & 16.74 & 57.27 & 12.34 & 40.02 & 17.37 & \textbf{25.80} & 16.47 & 0.686 & 0.514 \\
    \rowcolor{gray!12}
      & \methodname{}   & \textbf{35.11} & \textbf{64.27} & \textbf{25.21} & \textbf{55.36} & \textbf{18.86} & 22.60 & \textbf{19.61} & \textbf{0.828} & \textbf{0.277} \\
    \midrule
    \multirow{3}{*}{\texttt{6115eddb86}}
      & LangSplat       & 8.67 & 32.50 & 8.49 & 30.05 & 9.55 & 18.54 & 27.53 & 0.938 & 0.174 \\
      & OnlineLangSplat & 17.50 & 56.75 & 14.84 & 41.11 & 18.49 & 28.20 & 22.26 & 0.814 & 0.385 \\
    \rowcolor{gray!12}
      & \methodname{}   & \textbf{37.08} & \textbf{58.30} & \textbf{33.34} & \textbf{53.77} & \textbf{25.98} & \textbf{38.27} & \textbf{30.21} & \textbf{0.969} & \textbf{0.114} \\
    \midrule
    \multirow{3}{*}{\texttt{acd95847c5}}
      & LangSplat       & 7.42 & 19.34 & 8.54 & 19.67 & 9.47 & \textbf{18.96} & \textbf{30.13} & \textbf{0.973} & \textbf{0.107} \\
      & OnlineLangSplat & 7.88 & 30.89 & 6.75 & 32.87 & 9.83 & 14.84 & 24.49 & 0.825 & 0.407 \\
    \rowcolor{gray!12}
      & \methodname{}   & \textbf{20.14} & \textbf{39.47} & \textbf{22.82} & \textbf{35.91} & \textbf{12.15} & 15.68 & 28.09 & 0.963 & 0.132 \\
    \midrule
    \multirow{3}{*}{\texttt{bcd2436daf}}
      & LangSplat       & 11.34 & 32.83 & 18.00 & 44.38 & 12.11 & 21.02 & 31.81 & 0.960 & 0.116 \\
      & OnlineLangSplat & 5.94 & 46.19 & 21.34 & 51.80 & 20.20 & 27.53 & 25.99 & 0.862 & 0.316 \\
    \rowcolor{gray!12}
      & \methodname{}   & \textbf{35.28} & \textbf{60.64} & \textbf{39.68} & \textbf{63.40} & \textbf{25.68} & \textbf{32.14} & \textbf{32.58} & \textbf{0.967} & \textbf{0.113} \\
    \midrule
    \multirow{3}{*}{\texttt{c49a8c6cff}}
      & LangSplat       & 14.85 & 38.43 & 13.91 & 35.24 & 15.54 & \textbf{27.15} & \textbf{28.09} & \textbf{0.960} & \textbf{0.118} \\
      & OnlineLangSplat & 11.48 & 32.23 & 11.14 & \textbf{37.73} & 16.66 & 22.51 & 24.93 & 0.913 & 0.240 \\
    \rowcolor{gray!12}
      & \methodname{}   & \textbf{24.55} & \textbf{45.52} & \textbf{20.23} & 33.01 & \textbf{19.17} & 26.39 & 26.91 & 0.951 & 0.141 \\
    \midrule
    \multirow{3}{*}{\texttt{c50d2d1d42}}
      & LangSplat       & 12.61 & 44.48 & 15.42 & \textbf{53.05} & 17.49 & 34.93 & \textbf{30.69} & \textbf{0.973} & \textbf{0.108} \\
      & OnlineLangSplat & 12.54 & 56.48 & 9.42 & 40.68 & 13.51 & 21.89 & 23.83 & 0.818 & 0.406 \\
    \rowcolor{gray!12}
      & \methodname{}   & \textbf{27.83} & \textbf{61.23} & \textbf{32.24} & 51.50 & \textbf{22.68} & \textbf{37.03} & 28.73 & 0.961 & 0.127 \\
    \midrule
    \multirow{3}{*}{\texttt{fb5a96b1a2}}
      & LangSplat       & 17.68 & 44.45 & 20.84 & 54.13 & 12.83 & 22.45 & \textbf{26.91} & \textbf{0.958} & \textbf{0.145} \\
      & OnlineLangSplat & 10.16 & 46.81 & 5.17 & 55.22 & 14.77 & 23.09 & 22.78 & 0.864 & 0.325 \\
    \rowcolor{gray!12}
      & \methodname{}   & \textbf{33.46} & \textbf{59.10} & \textbf{36.07} & \textbf{64.34} & \textbf{22.54} & \textbf{35.29} & 25.83 & 0.945 & 0.169 \\
    \bottomrule
  \end{tabular}%
  }
  \label{tab:scannetpp_per_scene}
  \label{tab:scannetpp_per_scene_recon}
\end{table}

\paragraph{ScanNet++.} The exploration trajectories of \methodname{} are shown as green lines in Fig.~\ref{fig:trajectories}(b), and the text queries are listed in Table~\ref{tab:scannetpp_queries}. Table~\ref{tab:replica_per_scene} reports per-scene results for open-vocabulary semantic mapping and novel-view rendering. Our method achieves the highest 2D and 3D mIoU in all eight scenes and the highest localization accuracy on the SLAM views, demonstrating consistent semantic improvements across diverse real-world scenes. Rendering performance varies across scenes, with our method achieving the best PSNR, SSIM, and LPIPS in three of the eight scenes.

\subsection{Analysis of Image--Text Alignment in Semantic Features}
\label{app:embedding_alignment}

\paragraph{Experimental setup.}

To further investigate the embedding misalignment challenge identified in the introduction, we examine whether image features align with the text embeddings of their corresponding semantic
categories. We analyze 768D OpenCLIP features in the
\texttt{office0} scene of the Replica dataset.
We use the 2,000 RGB frames along the SLAM trajectory and their
ground-truth semantic labels.
For each frame, we extract semantic image regions using these
labels and group the resulting sub-images by category.
We retain the ten most frequent categories for this analysis.
For each category, we compute a category-level image embedding
by averaging the OpenCLIP image embeddings of its sub-images.
We also encode the corresponding ground-truth category name
using the OpenCLIP text encoder.

\paragraph{Similarity matrices.}
Let $\mathcal{R}_c$ denote the collection of sub-images assigned
to category $c$, and let $f_{\mathrm{img}}$ and $f_{\mathrm{text}}$
denote the OpenCLIP image and text encoders, respectively.
The category-level image and text embeddings are
\begin{equation}
    \boldsymbol{\mu}_c
    = \frac{1}{|\mathcal{R}_c|}
      \sum_{r \in \mathcal{R}_c} f_{\mathrm{img}}(r),
    \qquad
    \mathbf{t}_c = f_{\mathrm{text}}(\ell_c),
\end{equation}
where $\ell_c$ is the category name.
We construct two similarity matrices:
\begin{equation}
    S^{\mathrm{II}}_{ij}
    = \operatorname{cos}(\boldsymbol{\mu}_i,\boldsymbol{\mu}_j),
    \qquad
    S^{\mathrm{IT}}_{ij}
    = \operatorname{cos}(\boldsymbol{\mu}_i,\mathbf{t}_j).
\end{equation}
The first measures similarity between category-level image
embeddings, while the second measures cross-modal similarity
between image embeddings and category names.
\begin{wrapfigure}{r}{0.48\textwidth}
    \centering
    \vspace{-2pt}
\setlength{\parskip}{0pt}
\setlength{\abovecaptionskip}{2pt}
\setlength{\belowcaptionskip}{0pt}
    \includegraphics[
        width=\linewidth,
        trim=0bp 7bp 0bp 3bp,
        clip
    ]{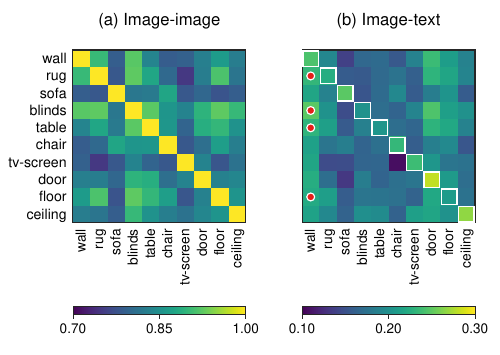}
    \vspace{-2pt}
    \caption{
\textbf{Category-level OpenCLIP similarity in Replica
\texttt{office0}.}
We aggregate image embeddings from ground-truth semantic
regions across 2,000 frames for the ten most frequent categories.
(a) Cosine similarity between category-averaged image embeddings.
(b) Cosine similarity between category-averaged image embeddings
(rows) and category-name text embeddings (columns).
White boxes indicate corresponding image--text category pairs;
red dots mark row maxima attained by an incorrect text category.
}
    \label{fig:embedding_mismatch}
    \vspace{-30pt}
\end{wrapfigure}

Fig.~\ref{fig:embedding_mismatch}(a) shows substantial
similarity between the average image embeddings of different
categories.
The smallest off-diagonal elements are greater than 0.7.
In the image--text matrix
(Fig.~\ref{fig:embedding_mismatch}(b)), the corresponding
category name does not always attain the highest similarity.
In particular, the average image embeddings of \textit{rug},
\textit{blinds}, \textit{table}, and \textit{floor} are more
similar to the text embedding of \textit{wall} than to their
corresponding category-name embeddings.

This analysis illustrates that category-averaged visual
features can remain ambiguous despite being embedded in a
language-aligned space.
Even when semantic regions are identified using ground-truth
labels, nearest-text matching does not consistently recover
their category names in the scene.
These results motivate careful treatment of semantic
distinctions when constructing language-queryable maps.
The analysis evaluates category-averaged features rather than individual regions and uses features before compression; it therefore does not account for additional effects introduced by feature compression or online adaptation.
\subsection{Semantic Preservation in Online Open-Vocabulary Mapping}
\label{app:supp_purity}

Active exploration exposes an online open-vocabulary mapper to an evolving and uneven distribution of observations. Newly encountered semantic content may initially occupy few pixels or appear in only a few views, while previously observed surfaces continue to dominate the training data. We investigate how feature compression can weaken semantic distinctions under these conditions, and how this motivates ActiveLang's prototype-balanced and separation-preserving adaptation, together with semantic supervision gating (Sec.~\ref{sec:semantic_mapping}).
We combine a stage-wise compression diagnostic with a comparison of final-map semantic purity on Replica room0.

\paragraph{Definition and evaluation protocol.}
We define \emph{purity} as point-wise semantic classification accuracy on the reconstructed Gaussian map. Ground truth labels are assigned from the nearest-neighbor points in the ground-truth point cloud (see below), while predictions are made
from the decoded language features stored in the map.
For this experiment, we retain Gaussians with opacity at least $0.1$ and subsample their centers using a $1$\,cm voxel grid. We construct the reference point cloud from the labeled ground truth mesh, applying the same sampling voxel size.

For a retained Gaussian center $\mathbf{x}_i$, let $\mathbf{z}_i$ denote its decoded, normalized language feature, and let $\mathbf{t}_c$ be the normalized text embedding of category $c$. The prediction for $\mathbf{x}_i$ is
\begin{equation}
    \hat y_i = \arg\max_{c\in\mathcal{C}}
    \mathbf{z}_i^{\mathsf T}\mathbf{t}_c.
\end{equation}
We conduct this experiment on Replica room0, which contains 23 semantic categories. All 23 categories satisfy the minimum support threshold of 50 points in the sampled reference cloud and are included in the candidate set $\mathcal{C}$.
Let $j(i)$ be the index of the nearest reference point, with position
$\mathbf{x}^{\mathrm{GT}}_{j(i)}$ and label $y^{\mathrm{GT}}_{j(i)}$.
We evaluate only spatially matched centers,
$\mathcal{M}=\{i:\|\mathbf{x}_i-\mathbf{x}^{\mathrm{GT}}_{j(i)}\|_2
\leq 0.05\,\mathrm{m}\}$, and compute
\begin{equation}
    \operatorname{Purity}
    = \frac{1}{|\mathcal{M}|}
      \sum_{i\in\mathcal{M}}
      \mathbb{1}\!\left[\hat y_i=y^{\mathrm{GT}}_{j(i)}\right].
    \label{eq:map_purity}
\end{equation}
Thus, purity measures semantic agreement on the spatially matched portion of the reconstructed map. It does not measure scene coverage, and centers farther than 5\,cm from the reference cloud are excluded instead of being counted as semantic errors.

\paragraph{Feature-mode imbalance and compression-induced collapse.}
The reconstruction loss is averaged over supervised pixels, so categories that occupy smaller image regions or appear in fewer observed views contribute less to the adaptation objective. Under this imbalance, minimizing average reconstruction error with a compact shared latent representation may fail to preserve the semantic distinctions needed to classify underrepresented categories correctly. For an underrepresented category, compression can alter feature directions enough that another category's text embedding becomes the nearest match. At the prediction level, this appears as \emph{semantic absorption}: content that was distinguishable before compression is decoded as another category. Here, underrepresentation is defined based on the online training distribution, rather than physical object size or final-map point count.

Table~\ref{tab:compression_chain} compares OnlineLangSplat and ActiveLang using compression diagnostics for the two-stage AE on Replica room0. For each method, sampled features are classified before compression (Raw), after being encoded-decoded by the frozen AE (AE), and after being encoded-decoded by the full AE+OLAE (Full). All classifications take place in the 768D decoded feature space. Samples and candidate labels are held fixed across stages within each run. The two runs report identical per-class sample counts and identical Raw and AE accuracies. They differ when the OLAE is invoked.

Compression changes semantic decisions unevenly. The frozen AE reduces lamp accuracy from 55.3\% to 0.7\%, table from 84.9\% to 21.5\%, and picture from 96.7\% to 27.3\%. Other categories survive this stage but are misclassified during online compression: ceiling falls from 87.5\% after AE to 0.0\% in both methods, with floor the most frequent prediction. OnlineLangSplat's sofa accuracy falls from 87.4\% to 24.0\%, with wall becoming its most frequent prediction, whereas floor and wall retain 99.9\% and 96.3\%. 
Thus, when weighted by pixel sampling frequency, the reconstruction loss can preserve frequently observed categories while shifting the decoded features of underrepresented categories toward those of more frequently observed categories.

ActiveLang improves Full-stage accuracy for the table category from 27.3\% to 55.2\% and picture from 63.8\% to 76.7\% relative to OnlineLangSplat. For table, the most frequent prediction changes from floor to the correct label. Window also improves from 31.7\% to 36.2\%, while cushion remains close at 89.1\% versus 89.4\%. These gains are selective: sofa falls to 0.0\% in ActiveLang, and lamp and ceiling remain at zero in both methods. Macro accuracy increases from 44.36\% to 45.74\%, a gain of 1.38 percentage points across the ten reported categories. 

These results indicate that compression is a source of classification errors and show partial recovery of particular semantic distinctions with ActiveLang. They are consistent with the feature-mode imbalance mechanism of Sec.~\ref{sec:semantic_mapping}.

\begin{table}[t]
\centering
\caption{\textbf{Compression-chain classification accuracy on Replica room0.}
Raw, AE, and Full denote classification in the original 768D feature space, after being encoded-decoded by the frozen AE, and after being encoded-decoded by the full AE+OLAE, respectively. Each method evaluates the same samples and candidate labels across stages. Both runs report identical sample counts and Raw/AE accuracies. Scores are percentages; $N$ is the diagnostic sample count, and Mode (category) is the most frequent Full-stage prediction. Bold marks the higher Full accuracy for each class across methods (ties are unbolded). Macro averages assign equal weight to the ten reported classes. 
}
\vspace{5pt}
\label{tab:compression_chain}
\begin{minipage}[t]{0.49\textwidth}\centering
\textbf{(a) OnlineLangSplat}\\[3pt]
\scriptsize
\setlength{\tabcolsep}{2pt}
\renewcommand{\arraystretch}{1.1}
\begin{tabular*}{\linewidth}{@{\extracolsep{\fill}}lrrrrl@{}}
\toprule
Class & $N$ & Raw & AE & Full & Mode \\
\midrule
cushion & 1,000 & 93.7 & 83.5 & 89.1 & cushion \\
floor & 1,000 & 94.0 & 99.3 & \textbf{99.9} & floor \\
lamp & 1,000 & 55.3 & 0.7 & 0.0 & wall \\
sofa & 1,000 & 91.9 & 87.4 & \textbf{24.0} & wall \\
wall & 1,000 & 93.9 & 95.7 & \textbf{96.3} & wall \\
window & 1,000 & 10.2 & 6.8 & 31.7 & wall \\
ceiling & 800 & 94.9 & 87.5 & 0.0 & floor \\
table & 776 & 84.9 & 21.5 & 27.3 & floor \\
rug & 686 & 67.5 & 46.6 & \textbf{11.5} & floor \\
picture & 600 & 96.7 & 27.3 & 63.8 & picture \\
\midrule
Macro & -- & 78.3 & 55.6 & 44.4 & -- \\
\bottomrule
\end{tabular*}
\end{minipage}
\hfill
\begin{minipage}[t]{0.49\textwidth}\centering
\textbf{(b) ActiveLang}\\[3pt]
\scriptsize
\setlength{\tabcolsep}{2pt}
\renewcommand{\arraystretch}{1.1}
\begin{tabular*}{\linewidth}{@{\extracolsep{\fill}}lrrrrl@{}}
\toprule
Class & $N$ & Raw & AE & Full & Mode \\
\midrule
cushion & 1,000 & 93.7 & 83.5 & \textbf{89.4} & cushion \\
floor & 1,000 & 94.0 & 99.3 & 96.8 & floor \\
lamp & 1,000 & 55.3 & 0.7 & 0.0 & wall \\
sofa & 1,000 & 91.9 & 87.4 & 0.0 & floor \\
wall & 1,000 & 93.9 & 95.7 & 93.6 & wall \\
window & 1,000 & 10.2 & 6.8 & \textbf{36.2} & wall \\
ceiling & 800 & 94.9 & 87.5 & 0.0 & floor \\
table & 776 & 84.9 & 21.5 & \textbf{55.2} & table \\
rug & 686 & 67.5 & 46.6 & 9.5 & floor \\
picture & 600 & 96.7 & 27.3 & \textbf{76.7} & picture \\
\midrule
Macro & -- & 78.3 & 55.6 & 45.7 & -- \\
\bottomrule
\end{tabular*}
\end{minipage}
\vspace{-15pt}
\end{table}

\paragraph{Connection to prototype-balanced and separation-preserving adaptation.}
The two-stage architecture in Sec.~\ref{sec:semantic_mapping} compresses 768D semantic features into frozen 32D codes and then into the 15D codes stored by the semantic Gaussians. Our contribution concerns how this online compression module learns from actively acquired observations. We use \emph{feature mode} to refer to a cluster of semantically similar embeddings in the frozen 32D feature space. Such a mode may capture a category or a particular appearance of that category, without requiring ground-truth labels. In Sec.~\ref{sec:semantic_mapping}, we address feature-mode imbalance with a bounded, class-agnostic prototype bank in the frozen 32D space. The balanced objective assigns equal weight to retained prototypes within its bank term, reducing the dependence of a mode's contribution on image-region area. Current-observation pixels complement this term by preserving within-mode variation. A prototype represents a feature mode, not a ground-truth category, so this mechanism does not require category labels during mapping.

Equal weighting alone does not constrain what happens to distinct modes after compression. The separation objective therefore penalizes excessive latent cosine similarity between prototypes that are sufficiently dissimilar in the frozen space. It complements balancing by explicitly discouraging compression-induced mode collapse. Both objectives operate on the OLAE; its encoded targets are detached when supervising the semantic Gaussians. This separates feature adaptation from map optimization and makes preservation of the semantic targets an explicit part of the mapping pipeline.

\paragraph{Semantic supervision gating.}
Active exploration can  produce observations dominated by a few feature modes or close-up views with insufficient semantic context. These observations can provide poor supervision even when the loss balances the retained prototypes. The gates in Sec.~\ref{sec:semantic_mapping} therefore control which observations update the OLAE and supervise the semantic Gaussians. Feature diversity governs eligibility for semantic updates, while the median-depth check excludes extreme close-up views. The forced OLAE update after repeated diversity-based rejections avoids prolonged adaptation pauses without bypassing the semantic-map gate (see Sec.~\ref{app:impl}).

\paragraph{Purity by Gaussian creation time.}
We also evaluated purity across different reconstruction stages using the final reconstructed Gaussian point cloud. We group sampled Gaussians using their creation timestamps and compute purity within each group. Group 1 contains the oldest Gaussians and Group 10 the newest. All groups are decoded separately using the final decoder: the curve therefore compares the purity of Gaussians across different creation-time; it does not record accuracy measured throughout training. The groups are computed separately for each method using their own decoder. In particular, active and predefined trajectories may observe different regions at the same relative stage, so corresponding deciles do not contain the same surfaces or semantic classes.

\begin{figure}[t]
    \centering
\includegraphics[width=0.75\textwidth,trim=0bp 0bp 0bp 33.6bp,clip]{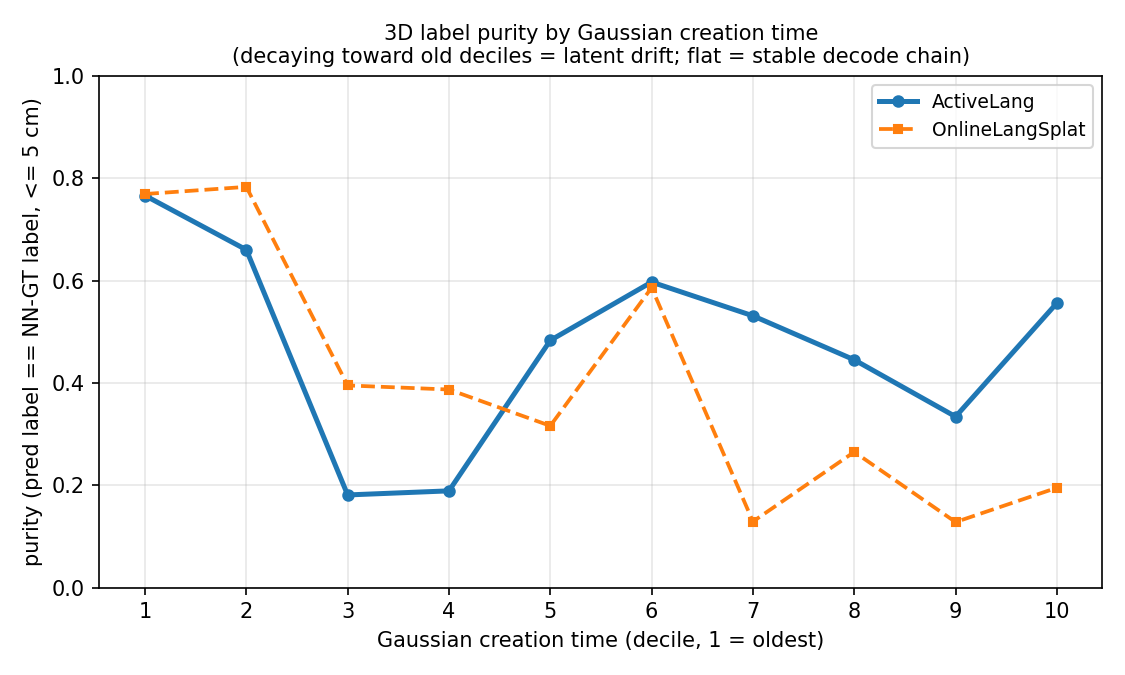}
    \vspace{-15pt}
    \caption{\textbf{Final-map semantic purity by Gaussian creation time on Replica room0.} Group 1 contains the oldest Gaussians and Group 10 the newest. All groups are evaluated at the final checkpoint using nearest-neighbor ground-truth labels within 5\,cm. ActiveLang shows higher purity than OnlineLangSplat in Groups 5-10, and its newest group exceeds the baseline. Groups are formed independently for each method and do not necessarily contain the same scene content, as different trajectories lead to different observations.. The curves describe the purity of Gaussians created in different stages, not longitudinal accuracy trajectories.}
    \label{fig:purity_age}
\end{figure}

OnlineLangSplat exhibits a decline toward later-created groups, despite a temporary increase in purity in Group 6. Purity is 76.9\% and 78.3\% in the first two cohorts, but only 12.9\%, 26.5\%, 12.8\%, and 19.5\% in cohorts 7--10. This pattern reveals low purity among many later-created Gaussians in the evaluated map. 

ActiveLang's profile is also non-monotonic: purity falls to 18.1\% and 18.9\% in Groups 3 and 4, then rises to 48.3\% and 59.7\% in Groups 5 and 6. It exceeds OnlineLangSplat in every group from 5 through 10. The newest group reaches 55.6\%, compared with 19.5\% for OnlineLangSplat. This supports improved semantic agreement in later-created portions of the map under the complete ActiveLang system.

These results suggest that compression can exacerbate semantic ambiguity, with sparsely observed categories becoming confused with background categories such as walls. Figure~\ref{fig:purity_age} further shows that ActiveLang's purity incrases for later-created groups after reaching low values in Groups 3-4, whereas OnlineLangSplat exhibits an overall decline toward later-created groups, 
ActiveLang achieves higher purity in every group from 5 through 10, suggesting better semantic preservation in newer portions of the map. 
These results are consistent with the intended role of ActiveLang's prototype bank in preserving underrepresented feature modes.

\subsection{Assets Used and Reproducibility}\label{app:reprod}

\subsubsection{Experimental result reproducibility}

We describe the full system in Sec.~\ref{sec:method} and the experimental protocol in Sec.~\ref{sec:exp}; per-scene results and the scene-specific text queries for both datasets are reported in Sec.~\ref{app:results} 
\methodname{} is built upon open-source software listed in Sec.~\ref{sec:assets}. More implementation details are listed in Sec.~\ref{app:impl}. Our code will be released upon acceptance.

\subsubsection{Compute resources for Experiments}
All experiments were performed on two NVIDIA RTX A6000 GPUs (48\,GB each). The first GPU hosts the dual 3DGS backbone: RGB-D Gaussian optimization, the gradient-based per-Gaussian language-field training (which shares the mapping loss), and the planner's rendering queries (geometric information gain and semantic uncertainty). The second GPU hosts the language-supervision pipeline that produces the per-pixel training targets: the dense CLIP encoder with its high-resolution refinement network, the general autoencoder, and the online adaptation of the scene-level autoencoder. The two memory footprints are strongly asymmetric: the language pipeline stays constant at ${\approx}6$\,GB throughout a run, while the mapping GPU grows with the map, from ${\approx}11$\,GB at initialization to ${\approx}25$\,GB after the exploration stages on Replica and peaking at ${\approx}35$\,GB depending on the scene size. This asymmetry motivates the two-GPU partition.

\subsubsection{Licenses for existing assets}\label{sec:assets}

\textbf{Datasets.}~~We conduct experiments on the following publicly available datasets. We list the URL, license, and citation for each dataset below.

\begin{enumerate}
    \item \textbf{Replica Dataset}~\citep{straub2019replica}
    \begin{itemize}
        \item URL: \url{https://github.com/facebookresearch/Replica-Dataset}
        \item License: Research or Education only (\url{https://github.com/facebookresearch/Replica-Dataset/blob/main/LICENSE})
    \end{itemize}
    \item \textbf{ScanNet++ Dataset}~\citep{yeshwanth2023scannetpp}
    \begin{itemize}
        \item URL: \url{https://kaldir.vc.in.tum.de/scannetpp/}
        \item License: Non-commercial research and education only (ScanNet++ Terms of Use)
    \end{itemize}
\end{enumerate}

\textbf{Software.}~~We use Habitat-Sim to render RGB-D observations from the dataset meshes. For mapping, we adopt SplaTAM, built on 3D Gaussian Splatting, as the backbone, and use a dense OpenCLIP extractor to produce pixel-aligned language features. We compare against LangSplat and Online Language Splatting using their released implementations. The source code for these components is available at:

\begin{enumerate}
    \item \textbf{Habitat-Sim}~\citep{savva2019habitat}
    \begin{itemize}
        \item URL: \url{https://github.com/facebookresearch/habitat-sim}
        \item License: MIT
    \end{itemize}

    \item \textbf{3D Gaussian Splatting (3DGS)}~\citep{kerbl20233dgs}
    \begin{itemize}
        \item URL: \url{https://github.com/graphdeco-inria/gaussian-splatting}
        \item License: Custom, non-commercial (\url{https://github.com/graphdeco-inria/gaussian-splatting/blob/main/LICENSE.md})
    \end{itemize}

    \item \textbf{SplaTAM}~\citep{keetha2024splatam}
    \begin{itemize}
        \item URL: \url{https://github.com/spla-tam/SplaTAM}
        \item License: BSD-3-Clause
    \end{itemize}

    \item \textbf{OpenCLIP}~\citep{cherti2023reproducible}
    \begin{itemize}
        \item URL: \url{https://github.com/mlfoundations/open_clip}
        \item License: MIT
    \end{itemize}

    \item \textbf{LangSplat}~\citep{qin2024langsplat}
    \begin{itemize}
        \item URL: \url{https://github.com/minghanqin/LangSplat}
        \item License: Gaussian-Splatting License, non-commercial research use (\url{https://github.com/minghanqin/LangSplat/blob/main/LICENSE.md})
    \end{itemize}

    \item \textbf{Online Language Splatting}~\citep{katragadda2025_onlinelang}
    \begin{itemize}
        \item URL: \url{https://github.com/rpng/online_lang_splatting}
        \item License: No repository-level license; the bundled 3DGS component is under the Gaussian-Splatting License
    \end{itemize}

    \item \textbf{Mip-Splatting}~\citep{yu2024mipsplatting}
    \begin{itemize}
        \item URL: \url{https://github.com/autonomousvision/mip-splatting.git}
        \item License: Gaussian-Splatting License, non-commercial research use (\url{https://github.com/autonomousvision/mip-splatting/blob/main/LICENSE.md})
    \end{itemize}

    \item \textbf{ActiveGAMER}~\citep{chen2025activegamer}
    \begin{itemize}
        \item URL: \url{https://github.com/oppo-us-research/ActiveGAMER.git}
        \item License: MIT
    \end{itemize}
    
\end{enumerate}

\subsection{Implementation Details}\label{app:impl}

\subsubsection{Simulation and Data}\label{sec:impl_data}

\paragraph{Rendering.}
All observations are rendered in Habitat-Sim~\citep{savva2019habitat} at $680\times1200$ resolution with a pinhole camera of focal length 600\,px, giving a $59.0^\circ$ vertical and $90.0^\circ$ horizontal field of view. Depth is stored at $1/6553.5$\,m per unit, and depth readings below 0.2\,m are treated as invalid.

\paragraph{Replica.}
We use 8 scenes: \texttt{office0}--\texttt{office4} and \texttt{room0}--\texttt{room2}. The passive baselines follow the standard 2,000-frame Replica trajectory~\citep{sucar2021imap}. We evaluate on (i) a 180-view novel-view orbit, with 36 azimuths $\times$ 5 heights on a 1.0\,m-radius circle facing inward, and (ii) every 10th frame of the SLAM trajectory (200 views). Fig.~\ref{fig:trajectories} (a) shows the trajectories under different settings overlaid on top-down views of the scenes.

\paragraph{ScanNet++.}
We use 8 scenes: \texttt{3f15a9266d}, \texttt{5eb31827b7}, \texttt{6115eddb86}, \texttt{acd95847c5}, \texttt{bcd2436daf}, \texttt{c49a8c6cff}, \texttt{c50d2d1d42}, and \texttt{fb5a96b1a2}.
We evaluate on (i) a 180-view inward-facing novel-view orbit whose radius is tuned per scene so that no view clips geometry, and (ii) every 20th frame real iPhone capture path (50 views). Trajectories under different settings are shown in Fig.~\ref{fig:trajectories} (b).

\subsubsection{Semantic Supervision Gates}
\label{app:semantic_gating}

Not every actively acquired observation provides suitable supervision for online semantic adaptation.
We therefore gate both OLAE updates and semantic-Gaussian supervision based on feature diversity and observation context.

For each frame, we greedily group up to 1,024 sampled and normalized 32D codes.
A code forms a new feature mode if its cosine similarity to all existing mode centers is below $0.6$.
After initialization, a frame is used to train the OLAE only if it contains at least eight distinct feature modes; on the Replica \texttt{room0} trajectory, typical observations contain 9--18 modes (median 12).
To avoid prolonged pauses in online adaptation, one OLAE update is forced after ten consecutive diversity-based rejections.

We additionally reject extreme close-up observations whose median valid depth is below $0.5$\,m, since such views may provide insufficient semantic context.
Semantic-Gaussian supervision also requires the feature-diversity criterion to be satisfied.
These gates affect only semantic learning, while RGB and depth reconstruction remain unaffected.

\subsubsection{Keyframe Selection}
\label{app:keyframe_selection}

We maintain three complementary types of observations for map refinement: local keyframes, global RGB-D keyframes, and global semantic keyframes.

\paragraph{Local Keyframes.}
Local keyframes form the 32-frame mapping window and support optimization of recently observed regions.

\paragraph{Global RGB-D Keyframes.}
Global RGB-D keyframes enable re-optimizing observations with poor geometric or appearance reconstruction.
Every 100 frames, excluding the five most recent observations, we add the lowest $30\%$ of previously unseen keyframes according to rendered-color PSNR.

\paragraph{Global Semantic Keyframes.}
Global semantic keyframes enable re-optimizing observations with high semantic reconstruction error.
Every 100 frames, we additionally add the lowest $30\%$ of previously unseen keyframes according to cosine similarity between rendered and extracted semantic features, computed at $1/4$ resolution.

The RGB-D and semantic criteria contribute to a shared global keyframe pool.
A keyframe's scores are recomputed only when its rendered content is likely to have changed, namely when at least 50 new Gaussians fall inside its frustum or when a newly mapped view overlaps its surface points by more than $5\%$.

\subsubsection{Exploration Planner}\label{sec:impl_planner}

\paragraph{Candidate view sampling.}
Following ActiveGAMER~\citep{chen2025activegamer}, we adopt a coarse-to-fine exploration strategy. We first use a coarse occupancy grid as exploration map to guide initial exploration, then densify the grid and resample unexplored voxels to guide the second exploration stage.
The exploration map is a 5\,cm voxel grid inside  the bounding box of the scene. Candidates are sampled from newly observed free space, with viewing directions drawn from a Fibonacci lattice on the sphere. Candidates closer than 0.5\,m to the nearest occupied voxel are discarded.

\paragraph{Scoring and termination.}
Candidates are scored with Eq.~\eqref{eq:next_best_view}, where the semantic-uncertainty term is enabled once the mean view coverage reaches 0.5. A candidate leaves the pool when its novel-pixel count falls below 0.5\% of the image. An exploration stage ends when the pool is empty, and a run ends after the last stage or at 2,000 steps. On ScanNet++, runs took 452--1,458 steps (889 on average), fewer than the 1,000 poses mapped by the passive baselines.

\paragraph{Motion and orientation.}
An RRT local planner advances up to 0.05\,m and $10^\circ$ per step. Every pose has zero roll relative to gravity. Transit poses face the next waypoint projected onto the horizontal plane and therefore also have zero pitch, while goal poses keep their sampled azimuth and pitch with roll removed.

\subsubsection{Baselines}\label{sec:impl_baselines}

All methods share the same RGB-D Guassian mapping backbone, ground-truth poses, Mip-Splatting filters, per-scene settings, and evaluation code. They differ only in their semantic pipelines and in whether a planner chooses the views.

\paragraph{OnlineLangSplat.}
We re-implementated Online Language Splatting~\citep{katragadda2025_onlinelang} on the shared backbone. It follows the passive trajectory and uses the same feature pipeline and OLAE schedule as ours, together with the dual-Gaussian representation and heteroscedastic loss without robust weighting. In place of the prototype bank, it uses a pixel replay buffer of 16 frames with 2 sampled per step. It uses neither supervision gates nor global keyframes.

\paragraph{LangSplat.}
A LangSplat~\citep{qin2024langsplat} pipeline on the shared backbone, following the passive trajectory. Masks are produced by SAM2 (\texttt{hiera-large}, $32\times32$ point grid, predicted-IoU threshold 0.7, stability threshold 0.85, minimum mask area 100\,px). Background-zeroed square crops are encoded by OpenCLIP~\citep{cherti2023reproducible} ConvNeXt-L (\texttt{laion2b\_s29b\_b131k\_ft\_soup}, 768-d, the same text space as ours). A scene-specific autoencoder compresses these features to 3 dimensions ($\ell_2+0.001\cos$ loss, Adam with learning rate $7\times10^{-4}$, 100 epochs) and is trained before mapping on every 10th frame of the passive trajectory. Each Gaussian stores a 3D feature supervised with an $\ell_1$ loss and shares the RGB-D geometry, without a separate semantic branch or uncertainty.

\subsubsection{Evaluation Protocol}\label{sec:impl_eval}

Each query is embedded with the OpenCLIP ConvNeXt-L text encoder, and its relevancy against the canonical negatives is $r=\min_{\mathrm{neg}}\operatorname{softmax}(10\,[s_{\mathrm{pos}},s_{\mathrm{neg}}])_0$~\citep{kerr2023lerf}.

\paragraph{2D segmentation and localization.}
Rendered features are decoded to 768 dimensions and scored per query. The relevancy map is smoothed with a $30\times30$ mean filter and averaged with the unsmoothed map, then min--max normalized and thresholded at 0.5 after mapping to $[-1,1]$ and clipping. A $7\times7$ majority filter follows, and IoU is computed against the ground-truth mask. Localization counts a hit when the argmax of the smoothed relevancy falls inside a ground-truth bounding box.

\paragraph{3D segmentation.}
The prediction consists of all semantic Gaussians with opacity $\geq0.1$, voxel-sampled at 1\,cm and labeled by relevancy $>0.5$. The ground truth is the full semantic mesh, restricted to the top-10 classes with at least 50 points each. Each ground-truth point receives the label of its nearest Gaussian, from which we compute mIoU and mAcc.

\paragraph{Geometry and rendering.}
RGB-D Gaussians with opacity $\geq0.1$, voxel-sampled at 1\,cm, are compared with uniform samples of the ground-truth mesh. We report accuracy and completion in cm as mean one-directional distances, and the completion ratio at a 5\,cm threshold. Rendering is evaluated with PSNR, SSIM, and LPIPS on RGB, and depth $\ell_1$ in cm over valid ground-truth depth.

\end{document}